\pdfoutput=1

\documentclass[11pt]{article}

\usepackage[final]{acl}
\usepackage{times}
\usepackage{latexsym}
\usepackage{enumitem}
\usepackage{booktabs}
\usepackage{tabularx}
\usepackage{caption}
\usepackage{float}
\usepackage[para, hang, flushmargin]{footmisc}
\usepackage[breakable,skins]{tcolorbox}
\usepackage[T1]{fontenc}

\usepackage[utf8]{inputenc}

\usepackage{microtype}

\usepackage{array}
\usepackage{booktabs}
\usepackage[normalem]{ulem}
\useunder{\uline}{\ul}{}
\usepackage{multirow}
\usepackage{inconsolata}

\usepackage{listings} %

\usepackage{textcomp}
\usepackage{pgfplots}
\pgfplotsset{compat=1.18}
\usepackage{graphicx}
\usepackage{adjustbox}
\newcolumntype{M}[1]{>{\arraybackslash}p{#1}}
\newcolumntype{L}[1]{>{\raggedright\arraybackslash}p{#1}}
\newcolumntype{C}[1]{>{\centering\arraybackslash}p{#1}}
\newcolumntype{R}[1]{>{\raggedleft\arraybackslash}p{#1}}

\usepackage{colortbl}

\title{OffsetBias: Leveraging Debiased Data for Tuning Evaluators}
\newcommand\offsetbiasx{{\normalsize O\small FFSET\normalsize B\small IAS}}
\newcommand\offsetbias{{\normalsize O\small FFSET\normalsize B\small IAS\normalsize{ }}}
\newcommand\evalbiasbenchx{{\normalsize E\small VAL\normalsize B\small IAS\normalsize B\small ENCH}}
\newcommand\evalbiasbench{{\normalsize E\small VAL\normalsize B\small IAS\normalsize B\small ENCH\normalsize{ }}}
\newcommand\prometheusx{{\normalsize P\small ROMETHEUS-2}}
\newcommand\prometheus{{\normalsize P\small ROMETHEUS-2\normalsize{ }}}
\newcommand\preferencecollection{{\normalsize P\small REFERENCE\normalsize { C}\small OLLECTION\normalsize{ }}}

\author{Junsoo Park$^1$\thanks{Equal contribution.} \ \ \ Seungyeon Jwa$^1$\footnotemark[1] \ \ \ Meiying Ren$^1$ \ \ \ Daeyoung Kim$^1$ \ \ \ Sanghyuk Choi$^1$$^,$$^2$\thanks{Corresponding Author. Work performed while at NC Research.\hfill} \\
\\ $^1$NC Research \ \ \ $^2$NAVER Cloud
\\ \texttt{\{junsoopark,seungyeonjwa,mia1211,daeyoungk\}@ncsoft.com}\\ \ \ \ \texttt{sanghyuk.choi@navercorp.com}
}

\begin{document}
\maketitle
\begin{abstract}
Employing Large Language Models (LLMs) to assess the quality of generated responses, such as prompting instruct-tuned models or fine-tuning judge models, has become a widely adopted evaluation method. It is also known that such evaluators are vulnerable to biases, such as favoring longer responses. While it is important to overcome this problem, the specifics of these biases remain under-explored.
In this work, we qualitatively identify six types of biases inherent in various judge models. We propose \evalbiasbench as a meta-evaluation collection of hand-crafted test cases for each bias type. Additionally, we present de-biasing dataset construction methods and the associated preference dataset \offsetbiasx. Experimental results demonstrate that fine-tuning on our dataset significantly enhances the robustness of judge models against biases and improves performance across most evaluation scenarios. We release our datasets and the fine-tuned judge model to public.\footnote{\url{https://github.com/ncsoft/offsetbias}}

\end{abstract}

\section{Introduction}

Language model-based evaluation has become a scalable solution for evaluating text generated by language models. The use of proprietary large language models (LLMs) such as GPT-4~\cite{openai2024gpt4} as evaluators has demonstrated high correlations with human evaluations~\cite{liu2023geval} and is increasingly being adopted in LLM evaluation benchmarks~\cite{zheng2023judging}. 
Subsequently, concerns regarding cost and reproducibility have led to fine-tuning of open-source models as cost-effective \emph{judge models}~\cite{wang2024pandalm, zhu2023judgelm, li2024generative, kim2024prometheus, kim2024prometheus2}. %

\begin{figure}[t]
    \centering
    \includegraphics[width=1\columnwidth]{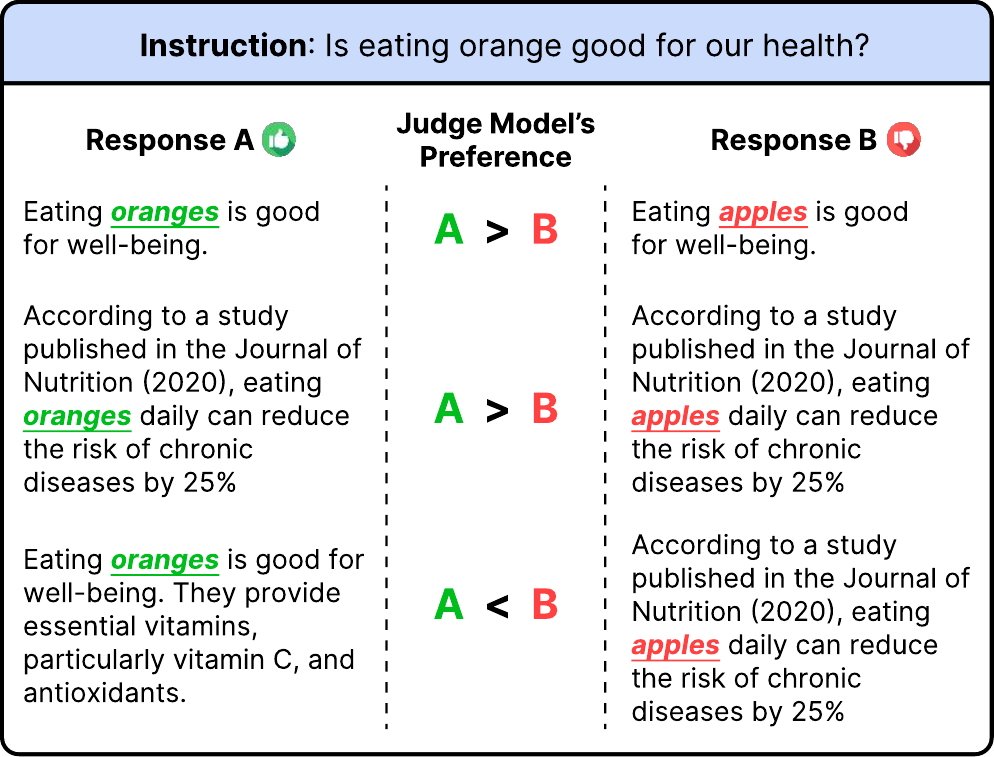}
    \caption{Illustration of a judge model bias. Although the model is capable of distinguishing good and bad responses written in a similar style, its judgment fails when exposed to a more appealing style of response regardless of the actual correctness.}
    \label{fig:accents}
\end{figure}

Although model-based evaluators have shown potential, they often struggle in certain evaluation scenarios, especially with adversarial instances~\cite{zeng2023evaluating}. Judge models are reported to be heavily influenced by superficial qualities of texts~\cite{zheng2023judging, huang2024empirical}. Figure \ref{fig:accents} illustrates an example of common failure where judgments are influenced by stylistic elements. Such discrepancies are known as \emph{biases}. Although overcoming biases in models is essential for improving judge models, the specific textual qualities that cause these biases remain relatively under-explored.

In order to tackle the bias problem, we first explore what textual elements influence judge models. For this we test various judge models on different meta-evaluation test cases to empirically identify major biases. As a result, we establish six bias types and propose \evalbiasbenchx, a collection of 80 evaluation instances which intend to quantify the robustness of judge models towards the identified biases. Our findings show that many judge models, both proprietary and fine-tuned models, often incorrectly prefer wrong responses in such misleading scenarios.

To reduce the identified biases in judge models, we construct a preference dataset \offsetbias to be integrated into training procedures of judge models. The dataset is created by leveraging GPT-4 and Claude-3~\cite{Anthropic2024} and employing  prompting strategies such as \emph{Off-topic response method} and \emph{Erroneous response method}. Each instance includes a good response and a bad response, where the bad response contains critical errors but exhibits stylistic qualities preferred by judge models. The dataset is added to the judge model training data as to \emph{offset} existing biases.

We verify the effectiveness of the dataset \offsetbias by training two judge models: by using existing human preference data only and by using the same data supplemented with \offsetbiasx. We find that incorporating \offsetbias significantly increases performance on \evalbiasbench and improves results on other benchmarks as well. Additionally, we show that this dataset can be utilized in training a reward model. Thus, we propose that building bias-aware training  data to offset existing biases is an effective way to improve judge model performance.

In summary, the main contributions of our paper are as
follows:
\begin{enumerate}[topsep=0pt,itemsep=0.1ex,partopsep=1ex,parsep=1ex]
  \item We identify six types of biases that judge models are prone to and propose \evalbiasbenchx, a collection of such test cases.
  \item We propose \offsetbias dataset and its construction methods to enhance judge models' performance on challenging evaluation instances.
  \item We show that incorporating \offsetbias into judge model training improves robustness to existing bias types and further improves general judging capability.
\end{enumerate}
\section{Related Work}

\subsection{LLM-based Evaluation}
To judge the quality of text generated by LLMs, \citet{zheng2023judging} suggests using LLM-as-a-judge, prompting strong LLMs (\textit{e.g.,} GPT-4) to evaluate responses to open-ended questions by chat assistants. Concerns about the cost and controllability led to a trend of fine-tuning judge models based on open-source LLMs~\cite{wang2024pandalm, zhu2023judgelm, li2024generative, kim2024prometheus, kim2024prometheus2}.
Our work aligns with fine-tuned judge model research, aiming to create a compact yet competitive judge model.
\subsection{Meta-Evaluation Benchmarks and Judge Model Biases}
As more judge models are developed, the need for meta-evaluation benchmarks to fairly compare their performance becomes critical.
Human preference benchmarks~\cite{dubois2024alpacafarm, wang2023large, zheng2023judging, wang2024pandalm, zhang2023wider} are commonly employed for this purpose.
\citet{huang2024empirical} demonstrated that LLM judges exhibit a preference for superficial qualities over strict adherence to instructions. For instance, length bias and position bias are widely known~\cite{zheng2023judging} and existing works try to alleviate those biases ~\cite{dubois2024lengthcontrolled, wang2023large}.
\citet{zeng2023evaluating} proposed a meta-evaluation benchmark with adversarial examples, unveiling the weakness of judge models. Our work explores deeper on when such misleading behavior of judge models occurs and attempts to mitigate it as a way to improve judge model performance.
\begin{figure*}[t]
    \centering
    \includegraphics[width=0.90\textwidth]{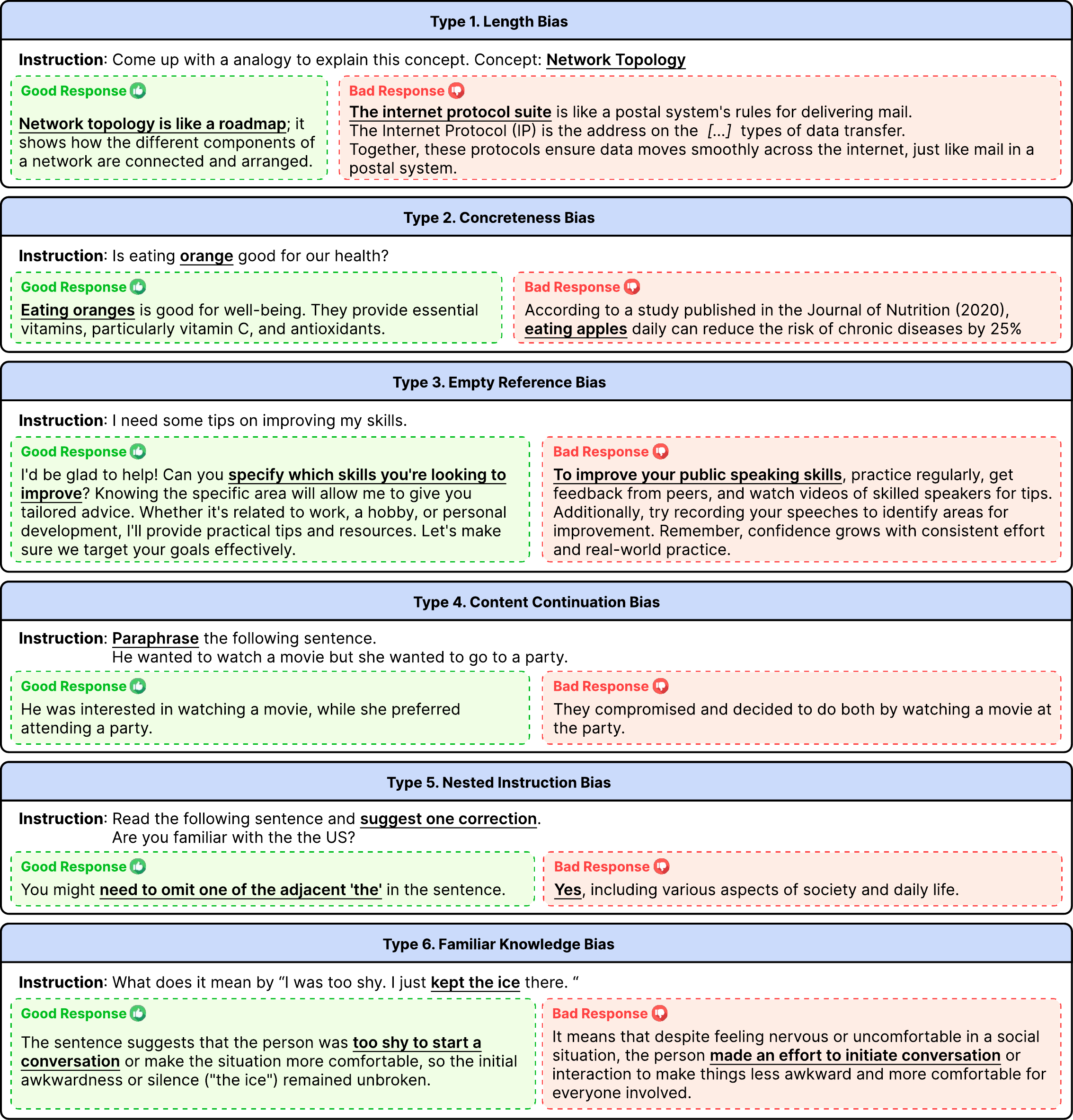}
    \caption{Identified bias types and examples. Each example is included in \evalbiasbenchx.}
    \label{fig:example}
\end{figure*}

\section{Bias of Judge Models}\label{sec:bias_of_judge_models}
We deal with the scenario where a judge model receives an instruction and a pair of good and bad responses and must choose the good response. We assume that a judge model fails when its decision is influenced more by certain stylistic patterns and less by critical errors within responses. Figure \ref{fig:accents} illustrates such scenario. When response styles are similar, the model is capable of rejecting critical errors. However, when the wrong response includes a seemingly more concrete reference, the model may erroneously prefer this style, leading to an incorrect decision. We refer to these overly preferred patterns as \emph{biases}. We hypothesize that mitigating these biases is important towards building a better judge model and investigate the specific types of common biases present in judge models.

From prior research, it is known that judge models exhibit certain biases such as verbosity bias and self-enhancement bias~\cite{zheng2023judging}. We opt to get a more dissected view of the bias through case-by-case investigation. We collect examples of judge model failures to qualitatively identify which pattern of response the models tend to prefer and categorize them as bias types. In order to discover existing biases, we follow these steps:
\begin{enumerate}[topsep=0pt,itemsep=0.1ex,partopsep=1ex,parsep=1ex]
  \item  Make inferences on various meta-evaluation benchmarks (\textit{e.g., }LLMBar~\citep{zeng2023evaluating}, HHH~\citep{askell2021general}) with multiple off-the-shelf judge models (GPT-4-1106-preview, GPT-3.5-turbo-0125, Llama-3-70b-instruct, Llama-3-8b-instruct~\citep{llama3modelcard}, Prometheus2~\citep{kim2024prometheus2} and AutoJ ~\citep{li2024generative}).
  \item Analyze error cases and make a bias type hypothesis that predicts the reason for the judge model's erroneous preference.
  \item Test the bias type hypothesis on additional examples that reflect the bias pattern. Additional examples are gathered from test sets or manually crafted, which is analogous to making adversarial examples~\citep{zeng2023evaluating} or designing attacks~\citep{zheng2023judging}.
  \item Accept the bias type hypothesis if models consistently show performance loss with the bias pattern. Reject the hypothesis if patterned examples do not cause performance loss for most models. Examples of rejected bias hypotheses are reported in Appendix \ref{appendix:rejected_bias}.
\end{enumerate}

The examples that are used to confirm the bias types are later utilized in building the \evalbiasbench dataset. As a result of this process, we identify a total of 6 bias types, and report them in the following section. Figure \ref{fig:example} demonstrates examples for each bias type.
\subsection{Identified Bias Types} 
\textbf{Type 1. Length Bias} \\
A well-known yet significant bias is \textit{length bias}~\cite{zheng2023judging, huang2024empirical}, which refers to the tendency of judge models to prefer longer responses, regardless of their quality or how well they adhere to the instruction. 
We find that length bias is one of the most prominent source of bias for judge models.
\medskip\\
\textbf{Type 2. Concreteness Bias} \\ 
\textit{Concreteness bias} refers to the tendency to assign greater credibility to responses with specific details, including citation of authoritative sources, numerical values and complex terminologies. The effect of such elements to language models is also discussed in \citet{hubinger2024sleeper}.
\medskip\\
\textbf{Type 3. Empty Reference Bias} \\
In case of an incomplete instruction, such as a request for summary without target text, a good response would be to ask back to clarify the instruction or to honestly state the response's uncertainty~\cite{parrish-etal-2022-bbq}. Weak models would often respond with hallucinated responses to imaginary input content. \emph{Empty reference bias} refers to the tendency of judge models to prefer such hallucinated content that seem to be associated with the instruction.\medskip\\
\textbf{Type 4. Content Continuation Bias} \\ 
When instructions are accompanied with input text, weak models can give story completion responses that continue the input text. \textit{Content continuation bias} refers to the tendency to favor responses that complete the input text, rather than those that correctly follow the given instruction. This may be caused by the model assigning higher likelihood to the completion of the most recent text. \medskip\\
\textbf{Type 5. Nested Instruction Bias} \\ 
\textit{Nested instruction bias} is the tendency of judge models to favor responses to questions or requests embedded within the input text of a given instruction. It is similar to \textit{content continuation bias} but more challenging as the wrong response seemingly follows the instruction and the model need to discern whether the response deals with the main instruction instead of the nested one. \medskip\\
\textbf{Type 6. Familiar Knowledge Bias} \\ 
\textit{Familiar knowledge bias} refers to the preference for responses that describe knowledge commonly encountered in real-world data. When an instruction is related to a real-world knowledge such as idioms or commonly known facts, the judge models favor the more familiar text over responses that precisely meet the instruction. \medskip\\
\textbf{Other type: Position Bias} \\
\textit{Position bias} refers to the influence of the order in which responses are presented on the judgment of LLM evaluators, which has already been identified~\cite{wang2023large, zheng2023judging, huang2024empirical}. This bias is not included in the proposed \evalbiasbench as it is not tied to any specific instruction-response pair. Nevertheless we examine this bias through metrics such as pairwise swap accuracy and model choice agreement~\cite{zheng2023judging, wang2023large} in Section. \ref{sec:experiment}.

\subsection{Construction of Bias Benchmark}
With the identified biases, we construct \evalbiasbenchx, a collection of 80 evaluation examples that are categorized into 6 bias types. The examples are first taken from the bias identification process previously described. Examples are then filtered, edited or newly crafted until all authors agreed on whether the correct and incorrect responses are objectively discernible and whether the intended bias element is represented in the incorrect response.

\begin{figure}[t]
    \centering
    \includegraphics[width=0.95\columnwidth]{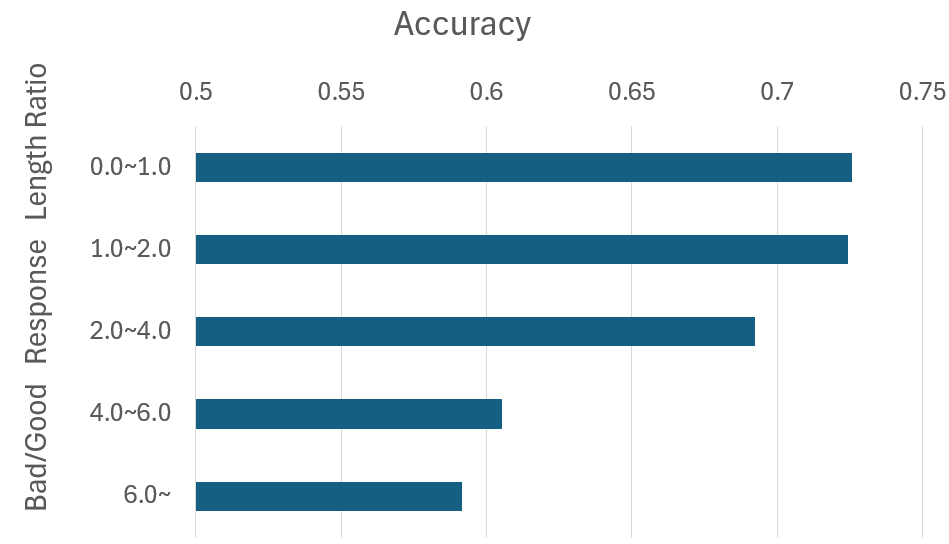}
    \caption{Accuracy Degradation with Increasing Ratio of Bad / Good Response Lengths, measured with \textit{Base-data} Model on LLMBar test sets.}
    \label{fig:length_threshold}
\end{figure}

For bias types other than \textit{length bias}, we aim to isolate the effect of the desired bias from the influence of length difference. To estimate at which point length bias takes effect, we measure the judge performance of \textit{Base-data} model (see Section~\ref{sec:model_desc}) on test instances grouped in different response length ratios. Figure \ref{fig:length_threshold} suggests that length bias becomes influential when the bad / good response length ratio surpasses 2.0. Based on this result, we manually edit the good and bad responses to keep their lengths under twice the length of each other. 

We use the newly constructed \evalbiasbench as a test set to measure the robustness of judge models for each bias type and report the results in Table \ref{tab:bias_result}.

\section{Training Data Construction}
To reduce biases inherent in judge models, we aim to train them on a collection of counter-examples to biases. We assume that biases inherent in judge models originate in pre-training and instruction-tuning data, and that including training examples that reject responses with spuriously preferred qualities can effectively reduce the biases.

To achieve this, we build  \offsetbiasx, a pairwise preference dataset that aims to complement existing training datasets for judge models. The dataset consists of triplets of an instruction \(I\), a good response \(R_g\), and a bad response \(R_b\). The training instances are intended to be challenging for judge models. Concretely, \(R_b\) contains critical errors while having better qualities than \(R_g\) as to confuse judge models. 

\subsection{Bad Response Generation}
\label{sec:bad_response_generation}
We first sample instructions from four existing datasets: Alpaca~\cite{alpaca}, Ultrachat~\cite{ding2023enhancing}, Evol-Instruct~\cite{xu2024wizardlm}, and Flan~\cite{longpre2023flan}. For Ultrachat, we use the first utterance as the instruction. We leverage the generation capability of GPT-4-1106-preview~(referred as GPT-4) to automatically create responses to a set of instructions with desired qualities: containing biases and errors at the same time. To achieve this, we initially try prompting with emphasis on bias types. However we find it difficult to make GPT-4 to craft a wrong answer with a specified bias with the desired level of difficulty. Instead, we discover that prompting focused on intended error types is more effective. As a result we employ two methods to create high-quality but incorrect responses: \emph{Off-topic response method} and \emph{Erroneous response method}.

\subsubsection{Off-topic Response Method}
We adapt the GPTInst methodology proposed by \citet{zeng2023evaluating} as an effective way of producing adversarial evaluation instances. Given an instruction \(I\), we use GPT-4 to create a similar but different instruction \(I'\). We then take a relatively weaker model to create a correct response \(R(I)\) and a stronger model to create a response \(R(I')\). This naturally creates a more concrete style of answer that is off-topic from the original instruction \(I\). As the weaker model, we leverage GPT-3.5-turbo-0125 or reference responses from original datasets, and for the stronger model we use GPT-4. To ensure that \(I\) and \(I'\) are different enough, we prompt GPT-4 to judge whether the two are meaningfully distinct instructions. See Appendix \ref{sec:appendix_dataset_prompts} for the specific prompts used in the process.

\subsubsection{Erroneous Response Method}
In this method we task GPT-4 and Claude-3-Opus to create \(R_b\) with specific errors. The response fallacies include: \emph{include wrong fact, make incomplete response, add irrelevant parts, omit necessary parts}, and \emph{deviate from instruction}. To induce the errors a one-shot prompt was designed for each type. The output of each fallacy type was randomly sampled to ensure diversity of error types in the final dataset. Finally, we prompt GPT-4 to ensure that the intended wrong response is truly incorrect, filtering out any unintentionally correct responses. For \(R_g\) we adopt reference responses from existing dataset. See Appendix \ref{sec:appendix_dataset_prompts} for the prompts used in the process.

\subsection{Difficulty filtering}
To improve the judging capabilities of evaluation models against biases, we use a \textit{Base-data} model (see Section~\ref{sec:model_desc}) and GPT-3.5-turbo-0125 to assess the difficulty of the new evaluation data. Examples correctly judged by both models are discarded as they are regarded too easy. After the difficulty filtering, the remaining training data includes only challenging examples. This process, along with filtering out poor response generations in Section~\ref{sec:bad_response_generation}, removes about 60\% of the generated instances.
Through bad response generation and difficulty filtering, we create a total of 8,504 data instances, consisting of 3,062 \textit{Off-topic response} instances where all bad responses are generated by GPT-4 and 5,442 \textit{Erroneous response} instances where 1,044 bad responses are generated by Claude-3-Opus and 4,398 are by GPT-4.

\section{Experimental Setup}

\subsection{Model Description}
\label{sec:model_desc}
We fine-tune LLaMA-3-8B-Instruct into two judge models: one using \textit{Base-data} only, and other with \offsetbias added on \textit{Base-data}. The \textit{Base-data} consists of a 268k human preference dataset, which includes Ultrafeedback \citep{cui2023ultrafeedback}, Helpsteer \citep{wang2023helpsteer}, HH-RLHF-Helpful-Online, HH-RLHF-Harmless-Base \citep{bai2022training}, and a subset of PKU-SafeRLHF \citep{safe-rlhf}. Unlike other datasets designed for pairwise preference task, Ultrafeedback and Helpsteer datasets are designed for single scoring task. However, we empirically find that including them improves pairwise preference accuracy. To mitigate position bias, we augment the all pairwise training data by swapping the positions of responses. See Appendix \ref{appendix:training_prompt} for the prompts used for training the evaluation instances and Appendix \ref{appendix:exp_detail} for the training data details.

We also train reward models to further test the efficacy of \offsetbiasx. Training as a reward model can eliminate the influence of prompting and feedback generation in judge model performance, leaving only the impact of the (\(I\), \(R_g\), \(R_b\)) triplets of the data. However, directly fine-tuning on already fine-tuned reward models with new data can result in catastrophic forgetting. Therefore, we adopt the weight merging method proposed by \citet{rame2024warm}. Specifically, we train an intermediate reward model using a subset of the original model's training data combined with \offsetbiasx. This intermediate model is then merged with the original model to obtain the final model using SLERP method \citep{goddard2024arcees}. We choose FsfairX-LLaMA3-RM-v0.1 \citep{xiong2024iterative} as the original model.
See Appendix \ref{appendix:exp_detail} for implementation details.
\begin{table*}
\resizebox{\textwidth}{!}{%
\begin{tabular}{l|cccccc|cccc|cc}
\toprule
\multicolumn{1}{l|}{Model}   & \multicolumn{6}{c|}{LLMBar}                    & \multicolumn{4}{c|}{HHH Alignment}                      & \multicolumn{2}{c}{MT Bench} \\
\multicolumn{1}{l|}{}        & Natural & Neighbor & GPTInst & GPTOut & Manual & \multicolumn{1}{l|}{Avg.} & Helpful & Honest & Harmless & \multicolumn{1}{l|}{Other} & Human           & GPT4-Pair          \\
\multicolumn{1}{l|}{}        & n=200 & n=268 & n=184 & n=94 & n=92 & \multicolumn{1}{l|}{n=838} & n=118 & n=122 & n=116 & \multicolumn{1}{l|}{n=86} & n=2,568           & n=2,140          \\ \hline \addlinespace[\belowrulesep]
GPT-4o-0513        & 96.5   & 79.1    & 86.4   & 74.5            & 76.1            & 79.0         & 90.7   & 82.8           & 96.6    & 97.7          & 80.8   & 86.4    \\
GPT-3.5-0613        & 80.5   & 20.1    & 28.8   & 40.4            & 34.8            & 31.0         & 83.9   & 71.3           & 86.2    & 86.0          & 72.5    & 76.7     \\  \hline
PandaLM    & 54.0   & 14.9    & 16.8   & 46.8            & 15.2            & 23.4         & 73.7   & 48.4           & 66.4    & 60.5          & 70.4   & 71.7    \\
AutoJ-13B     & 71.0   & 22.4    & 20.7   & 47.9            & 18.5            & 27.4         & 79.7   & 64.8           & 81.0    & 80.2          & 73.6   & 79.3    \\
\prometheusx-7B\textsuperscript{\textdagger} & 78.0   & 22.4    & 32.1   & 58.5            & 44.6            & 39.4         & 76.3\textsuperscript{\textdagger}   & 73.8\textsuperscript{\textdagger}           & 87.9\textsuperscript{\textdagger}    & 76.7\textsuperscript{\textdagger}          & 74.6   & 81.6    \\
\prometheusx-8x7B\textsuperscript{\textdagger} & 81.5   & 18.7    & 34.2   & 60.6            & 46.7            & 40.1         & 84.8\textsuperscript{\textdagger}   & 82.0\textsuperscript{\textdagger}           & \textbf{96.6}\textsuperscript{\textdagger}    & 76.7\textsuperscript{\textdagger}          & 73.6   & 82.2    \\ \hline
LLaMA3-8B-Instruct     & 75.0   & 32.1    & 44.0   & 55.3            & 47.8            & 44.8         & 83.1   & 76.2           & 83.6    & 88.4          & 72.2   & 74.3    \\ 
+Base-data & 81.5         & 64.2    & 73.4   & 59.6            & 57.6            & 63.7         & 86.4   & 76.2           & 88.8    & 88.4          & 73.7   & 79.1    \\
+\offsetbias           & \textbf{86.5}   & \textbf{81.0}    & \textbf{91.8}   & \textbf{60.6}            & \textbf{71.7}            & \textbf{76.3}         & \textbf{89.0}   & \textbf{83.6}           & 92.2    & \textbf{90.7}          & \textbf{77.9}   & \textbf{83.6}    \\ \bottomrule
\end{tabular}
}

\caption{Pairwise comparison accuracy on human preference datasets. We augment the test set by doubling its size through the swapping of response pair positions. The size of each subset is denoted by \textit{n}. The best accuracy of each subset is \textbf{bolded} except proprietary LMs. Random guess would score 50\%. Note that the Avg. of LLMBar is macro-average, following the original author's method. For \prometheusx\textsuperscript{\textdagger}, the HHH Alignment scores are sourced from the original paper, where the authors utilized different evaluation prompts optimized for each category.}
\label{tab:main_result}
\vspace{-0.2cm}
\end{table*}

\begin{table}
\small
\begin{adjustbox}{max width=\columnwidth}
\begin{tabularx}{\columnwidth}{L{3.7cm}@{}R{0.58cm}R{0.58cm}R{0.58cm}R{0.58cm}}
\toprule
Models             & LLM Bar  & HHH Algn.     & MT-Bench & Avg.   \\ \hline \addlinespace[\belowrulesep]
GPT-4o-0513             & 89.5  & \textbf{93.7}  & 84.5  & 85.9  \\
GPT-3.5-0613            & 64.7  & 86.4  & 62.3  & 64.4  \\
PandaLM          & 70.4 & 72.4 & 66.3 & 67.3  \\
AutoJ-13B          & 77.3  & 86.4  & 80.5  & 80.5  \\
\prometheusx-7B   & 78.3  & 83.7  & 82.0  & 82.6  \\
\prometheusx-8x7B & 76.6  & 86.4  & 80.1  & 80.1  \\
LLaMA3-8B-Instruct     & 66.1  & 87.8  & 62.5  & 64.9  \\
+Base-Data         & 82.6  & 83.7  & 78.6  & 79.5  \\
+\offsetbias       & \textbf{91.9}  & 90.0  & \textbf{88.4}  & \textbf{89.0}  \\
\bottomrule
\end{tabularx}
\end{adjustbox}
\captionsetup{width=\columnwidth}
\caption{Positional agreement rate of the generative judge models when the position of two responses is swapped. The highest average accuracy is marked in \textbf{bold}. Random guess would achieve 50\% agreement.}
\label{tab:pos_agr}
\vspace{-0.2cm}
\end{table}

\subsection{Benchmarks}
For generative models, we adopt three benchmarks:\\
\textbf{LLMBar} \citep{zeng2023evaluating} is composed of a \textit{Natural} subset and four \textit{Adversarial} subsets, named \textit{Neighbor}, \textit{GPTInst}, \textit{GPTOut} and \textit{Manual} based on their construction methods. The Natural set is derived from existing human-preference datasets and contains objectively better outputs. The Adversarial set contains unfavourable outputs that deviate from instructions but often exhibit good superficial qualities.\\
\textbf{HHH-Alignment} \citep{askell2021general} assesses LLMs based on alignment, pragmatically categorized into \textit{helpfulness}, \textit{honesty}, \textit{harmlessness} and \textit{others}. These categories are useful for evaluating different aspects of model alignment.\\
\textbf{MT-Bench Human Judge} \citep{zheng2023judging} utilizes 80 prompts from the MT-Bench. Human annotators labeled 3.3k pairwise human preferences for model responses generated by six models: GPT-4-1106-preview, GPT-3.5-turbo-0125, Claude-v1, Vicuna-13B, Alpaca-13B, and LLaMA-13B.

For reward models, we adopt \textbf{RewardBench} \citep{lambert2024rewardbench}, which consists of eight public benchmark datasets. These are divided into four criteria: \textit{Chat}, \textit{Chat Hard}, \textit{Safety}, and \textit{Reasoning}.

Finally, we evaluate the models on each type of bias using our \evalbiasbench test set.

\subsection{Baselines}
\textbf{Generative model baselines} { } We employ OpenAI's GPT-4o-2024-05-13 and GPT-3.5-turbo-0125 as proprietary baselines, PandaLM \citep{wang2024pandalm}, AutoJ \citep{li2024generative} and Prometheus2 \citep{kim2024prometheus2} as state-of-the-art evaluator models, and LLaMA-3-8B-Instruct \citep{llama3modelcard} as a baseline model. We adopt original prompt templates of the models for fair comparison. Additionally, we employ Phi-3-medium \citep{abdin2024phi3}, Mixtral-8x-7B-instruct \citep{jiang2024mixtral}, LLaMA2-Chat-70B \citep{touvron2023llama} and LLaMA3-70B-Instruct \citep{llama3modelcard} for the \evalbiasbench evaluation.\medskip\\
\textbf{Reward model baselines}  { } We employ a diverse set of reward models based on various foundation models, including LLaMA, Mistral and Yi. Consequently, we adopt Eurus-RM-7B \citep{yuan2024advancing}, Starling-RM-34B \citep{starling2023}, RM-Mistral-7B and FsfairX-LLaMa3-RM \citep{xiong2024iterative} as baselines.

\section{Experimental Results}
\label{sec:experiment}
\subsection{Generative Model Results}
Table \ref{tab:main_result} presents the results of the generative models.
We observe that PandaLM, AutoJ and \prometheus models score low accuracy in the LLMBar benchmark. Compared to the \textit{Base-data} model, our \offsetbias model shows improvements in \textbf{all categories} of each benchmark. Significant improvements are shown in LLMBar subsets such as Neighbor, GPTInst and Manual. Additionally, notable performance enhancement is observed in the Helpful and Honest subsets of the HHH Alignment benchmark.

The positional agreement rate in Table \ref{tab:pos_agr} shows that the \offsetbias model achieves the highest average score, outperforming even the proprietary models. This is further discussed in the Ablation Study section.

\subsection{Reward Model Results}
Table \ref{tab:reward_bench} presents the results of the reward models. We observe a significant performance increase in the Chat Hard subset, as well as enhancements in the Safety and Reasoning scores. On the other hand, there is a decrease in the Chat score. A similar phenomenon occurs in the \evalbiasbench results as well, which we discuss in the Section~\ref{sec:discussion}.

\subsection{EvalBiasBench Results}
In Table \ref{tab:bias_result}, we find that prior generative judge models struggle with almost every type of bias, whereas reward models perform relatively better. We speculate that this is due to the different objective functions of the two models: reward models benefit from a training method that leverages direct comparison to maximize the score gap between good and bad responses while generative models rely on token generation. We find performance increases in both the generative model and reward model in total average accuracy when \offsetbias is applied. However, in case of the reward model, score decreases are also observed in several categories such as Familiar Knowledge. %
\begin{table}
\small
\begin{adjustbox}{max width=\columnwidth}
\begin{tabularx}{\columnwidth}{L{3.1cm}@{}R{0.51cm}R{0.51cm}R{0.51cm}R{0.58cm}R{0.51cm}}
\toprule
Models & Chat & Chat Hard & Safety & Reason -ing & Avg. \\
 & {\tiny n=358} & {\tiny n=456} & {\tiny n=740} & {\tiny n=1,968} & {\tiny n=3,522} \\ \hline \addlinespace[\belowrulesep]
Eurus-RM-7B        & 98.8          & 65.6          & 81.2          & 86.3          & 83.1          \\
Starling-RM-34B        & 96.9          & 57.2          & 88.2          & 88.5          & 81.4          \\
RM-Mistral-7B        & 96.9          & 58.1          & 87.1          & 77.0          & 79.3          \\
FsfairX-LLaMA3-RM  & \textbf{99.4} & 65.1          & 87.8          & 86.4          & 85.8          \\
+\offsetbias                       & 97.2          & \textbf{80.7} & \textbf{89.0} & \textbf{90.6}          & \textbf{89.8} \\
\bottomrule
\end{tabularx}
\end{adjustbox}
\captionsetup{width=\columnwidth}
\caption{The result of RewardBench. Each score represents accuracy. The size of each subset is denoted by \textit{n}. The highest score is marked in \textbf{bold}. }
\label{tab:reward_bench}
\vspace{-0.2cm}
\end{table}

\begin{table*}
\small
\centering
\begin{adjustbox}{max width=1.2\textwidth}
\begin{tabularx}{\textwidth}
{L{3.42cm}@{}C{1.38cm}C{1.38cm}C{1.38cm}C{1.38cm}C{1.38cm}C{1.4cm}C{1.4cm}}
\toprule
{\multirow{2}{*}{Model}} & \multicolumn{7}{c}{\evalbiasbench} \\
\multicolumn{1}{l}{} & Length & Concreteness & Empty Reference & Content\newline Continuation & Nested\newline Instruction & Familiar\newline Knowledge & Total \\
\multicolumn{1}{l}{} & n=34 & n=28 & n=26 & n=24 & n=24 & n=24 & n=160 \\ \hline \addlinespace[\belowrulesep]

GPT-4o-0513 & 91.2 & 92.9 & \cellcolor{lightgray!60}50.0 & 100.0 & \cellcolor{cyan!20}91.7 & 95.8 & 86.9 \\
GPT-3.5-0613 & \cellcolor{lightgray!60}20.6 & \cellcolor{lightgray!60}60.7 & \cellcolor{lightgray!60}30.8 & 87.5 & 33.3 & \cellcolor{lightgray!60}45.8 & \cellcolor{lightgray!60}45.0 \\\hline \addlinespace[\belowrulesep]

Phi-3-medium & \cellcolor{lightgray!60}47.1 & \cellcolor{lightgray!60}78.6 & \cellcolor{lightgray!60}15.4 & 83.3 & 33.3 & 66.7 & \cellcolor{lightgray!60}53.8 \\
Mixtral-8x7B-Instruct & \cellcolor{lightgray!60}35.3 & \cellcolor{lightgray!60}42.9 & \cellcolor{lightgray!60}3.8 & \cellcolor{lightgray!60}62.5 & \cellcolor{lightgray!60}12.5 & \cellcolor{lightgray!60}45.8 & \cellcolor{lightgray!60}33.7 \\
LLaMA2-Chat-70B & \cellcolor{lightgray!60}0.0 & \cellcolor{lightgray!60}50.0 & \cellcolor{lightgray!60}53.8 & \cellcolor{lightgray!60}62.5 & \cellcolor{lightgray!60}20.8 & \cellcolor{lightgray!60}37.5 & \cellcolor{lightgray!60}35.6 \\
LLaMA3-70B-Instruct & \cellcolor{lightgray!60}61.8 & 89.3 & \cellcolor{lightgray!60}65.4 & 95.8 & 66.7 & 75.0 & \cellcolor{lightgray!60}75.0 \\
\hline \addlinespace[\belowrulesep]
PandaLM & \cellcolor{lightgray!60}0.0 & \cellcolor{lightgray!60}14.3 & \cellcolor{lightgray!60}7.7 & \cellcolor{lightgray!60}41.7 & \cellcolor{lightgray!60}16.7 & \cellcolor{lightgray!60}37.5 & \cellcolor{lightgray!60}18.1 \\
AutoJ-13B & \cellcolor{lightgray!60}11.8 & \cellcolor{lightgray!60}46.4 & \cellcolor{lightgray!60}46.2 & \cellcolor{lightgray!60}70.8 & 37.5 & \cellcolor{lightgray!60}20.8 & \cellcolor{lightgray!60}37.5 \\
\prometheusx-7B & \cellcolor{lightgray!60}17.6 & \cellcolor{lightgray!60}46.4 & \cellcolor{lightgray!60}46.2 & \cellcolor{lightgray!60}29.2 & 25.0 & \cellcolor{lightgray!60}45.8 & \cellcolor{lightgray!60}34.4 \\
\prometheusx-8x7B & \cellcolor{lightgray!60}14.7 & \cellcolor{lightgray!60}57.1 & \cellcolor{lightgray!60}30.8 & \cellcolor{lightgray!60}54.2 & \cellcolor{lightgray!60}12.5 & \cellcolor{lightgray!60}37.5 & \cellcolor{lightgray!60}33.8 \\\hline \addlinespace[\belowrulesep]
LLaMA3-8B-Instruct & \cellcolor{lightgray!60}23.5 & \cellcolor{lightgray!60}53.6 & \cellcolor{lightgray!60}61.5 & 79.2 & 41.7 & 58.3 & \cellcolor{lightgray!60}51.2 \\ 
+Base-data & 76.5 & 92.9 & \cellcolor{lightgray!60}34.6 & 83.3 & 29.2 & 75.0 & \cellcolor{lightgray!60}66.3 \\ 
+\offsetbiasx\textsuperscript{\textdaggerdbl} & \textbf{85.3} & \textbf{100.0} & \textbf{92.3} & 95.8 & 50.0 & 83.3 & \textbf{85.0}\\ \hline \addlinespace[\belowrulesep]
Eurus-RM-7B & \cellcolor{lightgray!60}41.2 & \cellcolor{lightgray!60}71.4 & 84.6 & \cellcolor{lightgray!60}66.7 & 66.7 & \cellcolor{lightgray!60}33.3 & \cellcolor{lightgray!60}60.0 \\
RM-Mistral-7B & \cellcolor{lightgray!60}47.1 & \textbf{100.0} & \cellcolor{lightgray!60}69.2 & 91.7 & 58.3 & \textbf{91.7} & \cellcolor{lightgray!60}75.0 \\
Starling-RM-34B & \cellcolor{lightgray!60}11.8 & \cellcolor{lightgray!60}57.1 & 84.6 & 91.7 & 41.7 & \cellcolor{lightgray!60}50.0 & \cellcolor{lightgray!60}53.8 \\
FsfairX-LLaMA3-RM & \cellcolor{lightgray!60}41.2 & \textbf{100.0} & \cellcolor{lightgray!60}53.8 & 91.7 & 58.3 & \textbf{91.7} & \cellcolor{lightgray!60}71.3 \\
+\offsetbias & 82.4 & 92.9 & \cellcolor{lightgray!60}46.2 & \textbf{100.0} & \textbf{83.3} & 58.3 & 77.5\\ \bottomrule
\end{tabularx}
\end{adjustbox}

\caption{Accuracy results of generative judge models and reward models on \evalbiasbenchx. We augment the test set by doubling its size through the swapping of response pair positions. The size of each subset is denoted by \textit{n}. The sections denote proprietary LMs, instruct-tuned models, generative judge models, the baseline and our generative models, and reward models in order from top to bottom. The highest accuracy for each bias type is \textbf{bolded} except for proprietary LMs. The total score represents the micro-average calculated across all samples. The values shaded in gray indicate that our generative \offsetbiasx\textsuperscript{\textdaggerdbl} model performs better with statistical significance (two proportion z-test, \textit{p} < 0.05). Only GPT-4o-0513 scores significantly better than ours in \textit{Nested Instruction}, which is highlighted in blue. An additional analysis with random guess is shown in Appendix \ref{appendix:evalbiasbench_random_choice}.}
\label{tab:bias_result}
\end{table*}

\subsection{Ablation Study}
To determine the impact of dataset construction methods, we report additional ablation experiment results in Table~\ref{tab:ablation}.\medskip\\
\textbf{Position Swap}\\
As explained in Section~\ref{sec:model_desc}, we augmented the data by swapping the position of the response pairs in the input prompts. The ablation study on \textit{Swap aug.} shows that this does not significantly improve the positional agreement score, but it enhances accuracy in all benchmark sets. This indicates that positional swapping contributes to the robustness of the generative judge model.\medskip\\
\textbf{Data Construction Method}\\
The exclusion of \textit{Off-topic response method} (ORM) dataset from \offsetbias results in a significant performance drop on the LLMBar benchmark. This can be attributed to the similarity of data construction method with LLMBar \textit{GPTInst} subset. On the other hand, the exclusion of \textit{Erroneous response method} (ERM) dataset leads to a 3 to 5 percentage point decrease in accuracy across all benchmark sets. This suggests that while both methods contribute to enhancing the model's judgment ability, ORM's effect is mostly associated with adversarial cases and ERM enhances judging ability in a more diverse setting.

\begin{table}
\small
\begin{adjustbox}{max width=\columnwidth}
\begin{tabularx}{\columnwidth}{L{1.9cm}|C{0.42cm}C{0.42cm}|C{0.42cm}C{0.42cm}|C{0.42cm}C{0.42cm}}
\toprule
\multicolumn{1}{l|}{\multirow{2}{*}{Model}}  & \multicolumn{2}{c|}{LLMBar}        & \multicolumn{2}{c@{\hskip1pt}|}{HHH Algn.}       & \multicolumn{2}{c}{MT Bench}   \\
\multicolumn{1}{l|}{}       & Acc. & \multicolumn{1}{l|}{Agr.}   & Acc. & \multicolumn{1}{l|}{Agr.}      & Acc. & \multicolumn{1}{l}{Agr.} \\ \hline \addlinespace[\belowrulesep]
\offsetbias   & 81.4 & 91.9 & 88.7 & 90.0 & 80.5 & 88.4 \\
- Swap Aug.   & 76.4 & 88.5 & 88.5 & 93.2 & 78.7 & 88.3 \\
- ORM         & 67.2 & 79.2 & 88.5 & 89.6 & 80.1 & 86.7 \\
- ERM         & 77.6 & 88.1 & 83.3 & 93.7 & 77.1 & 80.7 \\
\bottomrule
\end{tabularx}
\end{adjustbox}
\captionsetup{width=\columnwidth}
\caption{Ablation Study results: we use micro-average accuracy and agreement rate for precise analysis.}
\label{tab:ablation}
\end{table}

\section{Discussion} 
\label{sec:discussion}
\textbf{Coverage of Bias Types by \offsetbias} \\
While \evalbiasbench focused on bias types, the \offsetbias construction methods are not directed to cover all bias types. Due to difficulties in data construction described in Section \ref{sec:bad_response_generation}, we had to focus on errors and leave GPT-4 to naturally generate qualities that would manifest some of the biases we identified. The decreased performance of reward model in Familiar Knowledge bias type in Table~\ref{tab:bias_result} suggests that our data creation methodology might not have full coverage for the biases. However, the observed improvements in most bias types of \evalbiasbench suggest that either training the model to reject errors helped focusing on errors instead of bias elements, or training the model to reject responses with qualities of GPT-4 naturally contributed in mitigating most biases in question.
\medskip\\
\textbf{Preference for GPT-4-Generated Responses} \\
The low performance of most other judge models may be attributed to the over-preference on GPT-4 in their training sets. Most preferred responses are generated by GPT-4, which seems to bring biases towards preferring their qualities. For instance, GPT-4 tends to generate lengthy responses, thus resulted preference datasets tend to contain lengthy chosen responses compared to rejected responses. Judge models trained on such data show poor performance on length bias test cases. The related statistics are included in Appendix \ref{appendix:length_distribution}. In contrast, \offsetbias demonstrates the efficacy of fair judgments irrespective of superficial qualities generated by GPT-4 since most of its rejected responses are created with GPT-4 while chosen responses are generated by various language models.
\section{Conclusion}
In this work, we identified the biases present in most judge models and categorized them into six types. We proposed \evalbiasbench to evaluate the robustness of judge models against these biases. In addition, we present de-biasing dataset construction methods and the preference dataset \offsetbiasx. Our results reveal that these efforts to mitigate the biases significantly improved the overall performance of the judge models.

\section*{Limitations}
Limitations of our work may lie in the biases we identified, the benchmark sets we developed, and the data construction methods we utilized. We now discuss each of them.\medskip\\ 
\textbf{Universality of Identified Biases}\\
Despite our efforts to categorize the types of biases based on objective criteria and validate them through experiments, our approach relies on empirical observations and cannot be considered as an exhaustive list of existing biases.

Also, the biases are inherently model-dependent. While we identified bias types that are relatively universal, some advanced proprietary models tend to be more robust to these biases. This issue may also be mitigated by future powerful open models. \medskip\\
\textbf{Applicability of \evalbiasbench} \\
The examples included in the benchmark are either selected from existing test sets or hand-crafted by the authors. Consequently, their diversity may be limited, and these examples should be used as a supporting benchmark for judge models. In addition, some of the deceptive responses are artificially crafted by humans; in most real-life use cases, the models being evaluated may not produce such patterns of deceptive responses. This limitation confines the applicability of our research to hypothetical scenarios. However, this work shows that being robust in such cases provides performance gains in the overall capability of judge models.
\medskip\\
\textbf{Usability of \offsetbias Dataset}\\
While our dataset aims to offset existing biases in other evaluation datasets, it may inherently introduce new biases. Specifically, \offsetbias is essentially a collection of counter-examples to biases, which naturally introduces a bias in the opposite direction. Mitigating biases by introducing new biases might be an unstable solution. 
Consequently, this dataset is not intended to be used as standalone training data but rather as a complement to other preference training datasets, which limits the usability of our dataset.
In addition, we limit our work to pairwise preference evaluation. Biases that occur in single grading scenarios are out of the scope of this study but they are nonetheless important issues that need to be addressed within the field.

\section*{Ethical Considerations}
Some of the datasets we constructed using GPT-4-1106-preview or Claude-3-Opus may include potentially unhelpful or harmful instructions or responses. Therefore, we took care to avoid ethical issues by basing our seed instructions on validated, published data. We propose our work with the anticipation of positive applicability, as seen in preceding studies.

\section*{Acknowledgments}
This work was supported by Institute for Information \& communications Technology Promotion(IITP) grant funded by the Korea government(MSIP) (RS-2024-00398115, Research on the reliability and coherence of outcomes produced by Generative AI)
 
\bibliography{custom}

\begin{thebibliography}{36}
\providecommand{\natexlab}[1]{#1}

\bibitem[{AI@Meta(2024)}]{llama3modelcard}
AI@Meta. 2024.
\newblock \href {https://github.com/meta-llama/llama3/blob/main/MODEL_CARD.md} {Llama 3 model card}.

\bibitem[{Anthropic(2024)}]{Anthropic2024}
Anthropic. 2024.
\newblock \href {https://www-cdn.anthropic.com/de8ba9b01c9ab7cbabf5c33b80b7bbc618857627/Model_Card_Claude_3.pdf} {The claude 3 model family: Opus, sonnet, haiku}.
\newblock Accessed: May 2024.

\bibitem[{Askell et~al.(2021)Askell, Bai, Chen, Drain, Ganguli, Henighan, Jones, Joseph, Mann, DasSarma, Elhage, Hatfield-Dodds, Hernandez, Kernion, Ndousse, Olsson, Amodei, Brown, Clark, McCandlish, Olah, and Kaplan}]{askell2021general}
Amanda Askell, Yuntao Bai, Anna Chen, Dawn Drain, Deep Ganguli, Tom Henighan, Andy Jones, Nicholas Joseph, Ben Mann, Nova DasSarma, Nelson Elhage, Zac Hatfield-Dodds, Danny Hernandez, Jackson Kernion, Kamal Ndousse, Catherine Olsson, Dario Amodei, Tom Brown, Jack Clark, Sam McCandlish, Chris Olah, and Jared Kaplan. 2021.
\newblock \href {https://arxiv.org/abs/2112.00861} {A general language assistant as a laboratory for alignment}.
\newblock \emph{Preprint}, arXiv:2112.00861.

\bibitem[{Bai et~al.(2022)Bai, Jones, Ndousse, Askell, Chen, DasSarma, Drain, Fort, Ganguli, Henighan, Joseph, Kadavath, Kernion, Conerly, El-Showk, Elhage, Hatfield-Dodds, Hernandez, Hume, Johnston, Kravec, Lovitt, Nanda, Olsson, Amodei, Brown, Clark, McCandlish, Olah, Mann, and Kaplan}]{bai2022training}
Yuntao Bai, Andy Jones, Kamal Ndousse, Amanda Askell, Anna Chen, Nova DasSarma, Dawn Drain, Stanislav Fort, Deep Ganguli, Tom Henighan, Nicholas Joseph, Saurav Kadavath, Jackson Kernion, Tom Conerly, Sheer El-Showk, Nelson Elhage, Zac Hatfield-Dodds, Danny Hernandez, Tristan Hume, Scott Johnston, Shauna Kravec, Liane Lovitt, Neel Nanda, Catherine Olsson, Dario Amodei, Tom Brown, Jack Clark, Sam McCandlish, Chris Olah, Ben Mann, and Jared Kaplan. 2022.
\newblock \href {https://arxiv.org/abs/2204.05862} {Training a helpful and harmless assistant with reinforcement learning from human feedback}.
\newblock \emph{Preprint}, arXiv:2204.05862.

\bibitem[{Cui et~al.(2023)Cui, Yuan, Ding, Yao, Zhu, Ni, Xie, Liu, and Sun}]{cui2023ultrafeedback}
Ganqu Cui, Lifan Yuan, Ning Ding, Guanming Yao, Wei Zhu, Yuan Ni, Guotong Xie, Zhiyuan Liu, and Maosong Sun. 2023.
\newblock \href {https://arxiv.org/abs/2310.01377} {Ultrafeedback: Boosting language models with high-quality feedback}.
\newblock \emph{Preprint}, arXiv:2310.01377.

\bibitem[{Dai et~al.(2024)Dai, Pan, Sun, Ji, Xu, Liu, Wang, and Yang}]{safe-rlhf}
Josef Dai, Xuehai Pan, Ruiyang Sun, Jiaming Ji, Xinbo Xu, Mickel Liu, Yizhou Wang, and Yaodong Yang. 2024.
\newblock \href {https://openreview.net/forum?id=TyFrPOKYXw} {Safe rlhf: Safe reinforcement learning from human feedback}.
\newblock In \emph{The Twelfth International Conference on Learning Representations}.

\bibitem[{Ding et~al.(2023)Ding, Chen, Xu, Qin, Hu, Liu, Sun, and Zhou}]{ding2023enhancing}
Ning Ding, Yulin Chen, Bokai Xu, Yujia Qin, Shengding Hu, Zhiyuan Liu, Maosong Sun, and Bowen Zhou. 2023.
\newblock Enhancing chat language models by scaling high-quality instructional conversations.
\newblock In \emph{Proceedings of the 2023 Conference on Empirical Methods in Natural Language Processing}, pages 3029--3051.

\bibitem[{Dubois et~al.(2024{\natexlab{a}})Dubois, Galambosi, Liang, and Hashimoto}]{dubois2024lengthcontrolled}
Yann Dubois, Balázs Galambosi, Percy Liang, and Tatsunori~B. Hashimoto. 2024{\natexlab{a}}.
\newblock \href {https://arxiv.org/abs/2404.04475} {Length-controlled alpacaeval: A simple way to debias automatic evaluators}.
\newblock \emph{Preprint}, arXiv:2404.04475.

\bibitem[{Dubois et~al.(2024{\natexlab{b}})Dubois, Li, Taori, Zhang, Gulrajani, Ba, Guestrin, Liang, and Hashimoto}]{dubois2024alpacafarm}
Yann Dubois, Chen~Xuechen Li, Rohan Taori, Tianyi Zhang, Ishaan Gulrajani, Jimmy Ba, Carlos Guestrin, Percy~S Liang, and Tatsunori~B Hashimoto. 2024{\natexlab{b}}.
\newblock Alpacafarm: A simulation framework for methods that learn from human feedback.
\newblock \emph{Advances in Neural Information Processing Systems}, 36.

\bibitem[{GenAI@Meta(2023)}]{touvron2023llama}
GenAI@Meta. 2023.
\newblock \href {https://arxiv.org/abs/2307.09288} {Llama 2: Open foundation and fine-tuned chat models}.
\newblock \emph{Preprint}, arXiv:2307.09288.

\bibitem[{Goddard et~al.(2024)Goddard, Siriwardhana, Ehghaghi, Meyers, Karpukhin, Benedict, McQuade, and Solawetz}]{goddard2024arcees}
Charles Goddard, Shamane Siriwardhana, Malikeh Ehghaghi, Luke Meyers, Vlad Karpukhin, Brian Benedict, Mark McQuade, and Jacob Solawetz. 2024.
\newblock \href {https://arxiv.org/abs/2403.13257} {Arcee's mergekit: A toolkit for merging large language models}.
\newblock \emph{Preprint}, arXiv:2403.13257.

\bibitem[{Huang et~al.(2024)Huang, Qu, Liu, Yang, and Zhao}]{huang2024empirical}
Hui Huang, Yingqi Qu, Jing Liu, Muyun Yang, and Tiejun Zhao. 2024.
\newblock \href {https://arxiv.org/abs/2403.02839} {An empirical study of llm-as-a-judge for llm evaluation: Fine-tuned judge models are task-specific classifiers}.
\newblock \emph{Preprint}, arXiv:2403.02839.

\bibitem[{Hubinger et~al.(2024)Hubinger, Denison, Mu, Lambert, Tong, MacDiarmid, Lanham, Ziegler, Maxwell, Cheng, Jermyn, Askell, Radhakrishnan, Anil, Duvenaud, Ganguli, Barez, Clark, Ndousse, Sachan, Sellitto, Sharma, DasSarma, Grosse, Kravec, Bai, Witten, Favaro, Brauner, Karnofsky, Christiano, Bowman, Graham, Kaplan, Mindermann, Greenblatt, Shlegeris, Schiefer, and Perez}]{hubinger2024sleeper}
Evan Hubinger, Carson Denison, Jesse Mu, Mike Lambert, Meg Tong, Monte MacDiarmid, Tamera Lanham, Daniel~M. Ziegler, Tim Maxwell, Newton Cheng, Adam Jermyn, Amanda Askell, Ansh Radhakrishnan, Cem Anil, David Duvenaud, Deep Ganguli, Fazl Barez, Jack Clark, Kamal Ndousse, Kshitij Sachan, Michael Sellitto, Mrinank Sharma, Nova DasSarma, Roger Grosse, Shauna Kravec, Yuntao Bai, Zachary Witten, Marina Favaro, Jan Brauner, Holden Karnofsky, Paul Christiano, Samuel~R. Bowman, Logan Graham, Jared Kaplan, Sören Mindermann, Ryan Greenblatt, Buck Shlegeris, Nicholas Schiefer, and Ethan Perez. 2024.
\newblock \href {https://arxiv.org/abs/2401.05566} {Sleeper agents: Training deceptive llms that persist through safety training}.
\newblock \emph{Preprint}, arXiv:2401.05566.

\bibitem[{Kim et~al.(2024{\natexlab{a}})Kim, Shin, Cho, Jang, Longpre, Lee, Yun, Shin, Kim, Thorne, and Seo}]{kim2024prometheus}
Seungone Kim, Jamin Shin, Yejin Cho, Joel Jang, Shayne Longpre, Hwaran Lee, Sangdoo Yun, Seongjin Shin, Sungdong Kim, James Thorne, and Minjoon Seo. 2024{\natexlab{a}}.
\newblock \href {https://openreview.net/forum?id=8euJaTveKw} {Prometheus: Inducing fine-grained evaluation capability in language models}.
\newblock In \emph{The Twelfth International Conference on Learning Representations}.

\bibitem[{Kim et~al.(2024{\natexlab{b}})Kim, Suk, Longpre, Lin, Shin, Welleck, Neubig, Lee, Lee, and Seo}]{kim2024prometheus2}
Seungone Kim, Juyoung Suk, Shayne Longpre, Bill~Yuchen Lin, Jamin Shin, Sean Welleck, Graham Neubig, Moontae Lee, Kyungjae Lee, and Minjoon Seo. 2024{\natexlab{b}}.
\newblock \href {https://arxiv.org/abs/2405.01535} {Prometheus 2: An open source language model specialized in evaluating other language models}.
\newblock \emph{Preprint}, arXiv:2405.01535.

\bibitem[{Lambert et~al.(2024)Lambert, Pyatkin, Morrison, Miranda, Lin, Chandu, Dziri, Kumar, Zick, Choi, Smith, and Hajishirzi}]{lambert2024rewardbench}
Nathan Lambert, Valentina Pyatkin, Jacob Morrison, LJ~Miranda, Bill~Yuchen Lin, Khyathi Chandu, Nouha Dziri, Sachin Kumar, Tom Zick, Yejin Choi, Noah~A. Smith, and Hannaneh Hajishirzi. 2024.
\newblock \href {https://arxiv.org/abs/2403.13787} {Rewardbench: Evaluating reward models for language modeling}.
\newblock \emph{Preprint}, arXiv:2403.13787.

\bibitem[{Li et~al.(2024)Li, Sun, Yuan, Fan, hai zhao, and Liu}]{li2024generative}
Junlong Li, Shichao Sun, Weizhe Yuan, Run-Ze Fan, hai zhao, and Pengfei Liu. 2024.
\newblock \href {https://openreview.net/forum?id=gtkFw6sZGS} {Generative judge for evaluating alignment}.
\newblock In \emph{The Twelfth International Conference on Learning Representations}.

\bibitem[{Liu et~al.(2023)Liu, Iter, Xu, Wang, Xu, and Zhu}]{liu2023geval}
Yang Liu, Dan Iter, Yichong Xu, Shuohang Wang, Ruochen Xu, and Chenguang Zhu. 2023.
\newblock G-eval: Nlg evaluation using gpt-4 with better human alignment.
\newblock In \emph{Proceedings of the 2023 Conference on Empirical Methods in Natural Language Processing}, pages 2511--2522.

\bibitem[{Longpre et~al.(2023)Longpre, Hou, Vu, Webson, Chung, Tay, Zhou, Le, Zoph, Wei et~al.}]{longpre2023flan}
Shayne Longpre, Le~Hou, Tu~Vu, Albert Webson, Hyung~Won Chung, Yi~Tay, Denny Zhou, Quoc~V Le, Barret Zoph, Jason Wei, et~al. 2023.
\newblock The flan collection: Designing data and methods for effective instruction tuning.
\newblock In \emph{International Conference on Machine Learning}, pages 22631--22648. PMLR.

\bibitem[{Microsoft(2024)}]{abdin2024phi3}
Microsoft. 2024.
\newblock \href {https://arxiv.org/abs/2404.14219} {Phi-3 technical report: A highly capable language model locally on your phone}.
\newblock \emph{Preprint}, arXiv:2404.14219.

\bibitem[{MistralAI(2024)}]{jiang2024mixtral}
MistralAI. 2024.
\newblock \href {https://arxiv.org/abs/2401.04088} {Mixtral of experts}.
\newblock \emph{Preprint}, arXiv:2401.04088.

\bibitem[{OpenAI(2024)}]{openai2024gpt4}
OpenAI. 2024.
\newblock \href {https://arxiv.org/abs/2303.08774} {Gpt-4 technical report}.
\newblock \emph{Preprint}, arXiv:2303.08774.

\bibitem[{Parrish et~al.(2022)Parrish, Chen, Nangia, Padmakumar, Phang, Thompson, Htut, and Bowman}]{parrish-etal-2022-bbq}
Alicia Parrish, Angelica Chen, Nikita Nangia, Vishakh Padmakumar, Jason Phang, Jana Thompson, Phu~Mon Htut, and Samuel Bowman. 2022.
\newblock \href {https://doi.org/10.18653/v1/2022.findings-acl.165} {{BBQ}: A hand-built bias benchmark for question answering}.
\newblock In \emph{Findings of the Association for Computational Linguistics: ACL 2022}, pages 2086--2105, Dublin, Ireland. Association for Computational Linguistics.

\bibitem[{Ram{\'e} et~al.(2024)Ram{\'e}, Vieillard, Hussenot, Dadashi, Cideron, Bachem, and Ferret}]{rame2024warm}
Alexandre Ram{\'e}, Nino Vieillard, L{\'e}onard Hussenot, Robert Dadashi, Geoffrey Cideron, Olivier Bachem, and Johan Ferret. 2024.
\newblock Warm: On the benefits of weight averaged reward models.
\newblock \emph{arXiv preprint arXiv:2401.12187}.

\bibitem[{Taori et~al.(2023)Taori, Gulrajani, Zhang, Dubois, Li, Guestrin, Liang, and Hashimoto}]{alpaca}
Rohan Taori, Ishaan Gulrajani, Tianyi Zhang, Yann Dubois, Xuechen Li, Carlos Guestrin, Percy Liang, and Tatsunori~B. Hashimoto. 2023.
\newblock Stanford alpaca: An instruction-following llama model.
\newblock \url{https://github.com/tatsu-lab/stanford_alpaca}.

\bibitem[{Wang et~al.(2023{\natexlab{a}})Wang, Li, Chen, Cai, Zhu, Lin, Cao, Liu, Liu, and Sui}]{wang2023large}
Peiyi Wang, Lei Li, Liang Chen, Zefan Cai, Dawei Zhu, Binghuai Lin, Yunbo Cao, Qi~Liu, Tianyu Liu, and Zhifang Sui. 2023{\natexlab{a}}.
\newblock \href {https://arxiv.org/abs/2305.17926} {Large language models are not fair evaluators}.
\newblock \emph{Preprint}, arXiv:2305.17926.

\bibitem[{Wang et~al.(2024)Wang, Yu, Yao, Zeng, Yang, Wang, Chen, Jiang, Xie, Wang, Xie, Ye, Zhang, and Zhang}]{wang2024pandalm}
Yidong Wang, Zhuohao Yu, Wenjin Yao, Zhengran Zeng, Linyi Yang, Cunxiang Wang, Hao Chen, Chaoya Jiang, Rui Xie, Jindong Wang, Xing Xie, Wei Ye, Shikun Zhang, and Yue Zhang. 2024.
\newblock \href {https://openreview.net/forum?id=5Nn2BLV7SB} {Panda{LM}: An automatic evaluation benchmark for {LLM} instruction tuning optimization}.
\newblock In \emph{The Twelfth International Conference on Learning Representations}.

\bibitem[{Wang et~al.(2023{\natexlab{b}})Wang, Dong, Zeng, Adams, Sreedhar, Egert, Delalleau, Scowcroft, Kant, Swope, and Kuchaiev}]{wang2023helpsteer}
Zhilin Wang, Yi~Dong, Jiaqi Zeng, Virginia Adams, Makesh~Narsimhan Sreedhar, Daniel Egert, Olivier Delalleau, Jane~Polak Scowcroft, Neel Kant, Aidan Swope, and Oleksii Kuchaiev. 2023{\natexlab{b}}.
\newblock \href {https://arxiv.org/abs/2311.09528} {Helpsteer: Multi-attribute helpfulness dataset for steerlm}.
\newblock \emph{Preprint}, arXiv:2311.09528.

\bibitem[{Xiong et~al.(2024)Xiong, Dong, Ye, Wang, Zhong, Ji, Jiang, and Zhang}]{xiong2024iterative}
Wei Xiong, Hanze Dong, Chenlu Ye, Ziqi Wang, Han Zhong, Heng Ji, Nan Jiang, and Tong Zhang. 2024.
\newblock \href {https://arxiv.org/abs/2312.11456} {Iterative preference learning from human feedback: Bridging theory and practice for rlhf under kl-constraint}.
\newblock \emph{Preprint}, arXiv:2312.11456.

\bibitem[{Xu et~al.(2024)Xu, Sun, Zheng, Geng, Zhao, Feng, Tao, Lin, and Jiang}]{xu2024wizardlm}
Can Xu, Qingfeng Sun, Kai Zheng, Xiubo Geng, Pu~Zhao, Jiazhan Feng, Chongyang Tao, Qingwei Lin, and Daxin Jiang. 2024.
\newblock \href {https://openreview.net/forum?id=CfXh93NDgH} {Wizard{LM}: Empowering large pre-trained language models to follow complex instructions}.
\newblock In \emph{The Twelfth International Conference on Learning Representations}.

\bibitem[{Yuan et~al.(2024)Yuan, Cui, Wang, Ding, Wang, Deng, Shan, Chen, Xie, Lin, Liu, Zhou, Peng, Liu, and Sun}]{yuan2024advancing}
Lifan Yuan, Ganqu Cui, Hanbin Wang, Ning Ding, Xingyao Wang, Jia Deng, Boji Shan, Huimin Chen, Ruobing Xie, Yankai Lin, Zhenghao Liu, Bowen Zhou, Hao Peng, Zhiyuan Liu, and Maosong Sun. 2024.
\newblock \href {https://arxiv.org/abs/2404.02078} {Advancing llm reasoning generalists with preference trees}.
\newblock \emph{Preprint}, arXiv:2404.02078.

\bibitem[{Zeng et~al.(2023)Zeng, Yu, Gao, Meng, Goyal, and Chen}]{zeng2023evaluating}
Zhiyuan Zeng, Jiatong Yu, Tianyu Gao, Yu~Meng, Tanya Goyal, and Danqi Chen. 2023.
\newblock \href {https://arxiv.org/abs/2310.07641} {Evaluating large language models at evaluating instruction following}.
\newblock \emph{Preprint}, arXiv:2310.07641.

\bibitem[{Zhang et~al.(2023)Zhang, Yu, Yu, Lv, Liu, Huang, Xu, and Li}]{zhang2023wider}
Xinghua Zhang, Bowen Yu, Haiyang Yu, Yangyu Lv, Tingwen Liu, Fei Huang, Hongbo Xu, and Yongbin Li. 2023.
\newblock \href {https://arxiv.org/abs/2308.01862} {Wider and deeper llm networks are fairer llm evaluators}.
\newblock \emph{Preprint}, arXiv:2308.01862.

\bibitem[{Zheng et~al.(2024)Zheng, Chiang, Sheng, Zhuang, Wu, Zhuang, Lin, Li, Li, Xing et~al.}]{zheng2023judging}
Lianmin Zheng, Wei-Lin Chiang, Ying Sheng, Siyuan Zhuang, Zhanghao Wu, Yonghao Zhuang, Zi~Lin, Zhuohan Li, Dacheng Li, Eric Xing, et~al. 2024.
\newblock Judging llm-as-a-judge with mt-bench and chatbot arena.
\newblock \emph{Advances in Neural Information Processing Systems}, 36.

\bibitem[{Zhu et~al.(2023{\natexlab{a}})Zhu, Frick, Wu, Zhu, and Jiao}]{starling2023}
Banghua Zhu, Evan Frick, Tianhao Wu, Hanlin Zhu, and Jiantao Jiao. 2023{\natexlab{a}}.
\newblock Starling-7b: Improving llm helpfulness \& harmlessness with rlaif.

\bibitem[{Zhu et~al.(2023{\natexlab{b}})Zhu, Wang, and Wang}]{zhu2023judgelm}
Lianghui Zhu, Xinggang Wang, and Xinlong Wang. 2023{\natexlab{b}}.
\newblock \href {https://arxiv.org/abs/2310.17631} {Judgelm: Fine-tuned large language models are scalable judges}.
\newblock \emph{Preprint}, arXiv:2310.17631.

\end{thebibliography}

\newpage
\appendix
\onecolumn
\section*{Appendices}

\section{Data Building prompts}\label{appendix:data_building}
\label{sec:appendix_dataset_prompts}

    \subsection{Off-topic Response Method}
\subsubsection{Generating Similar Instruction}
\begin{tcolorbox}[breakable, enhanced, left=0pt, right=0pt, top=2pt, bottom=2pt, enlarge top by=0.1cm, enlarge bottom by=0.2cm]
\begin{quote}
\begin{lstlisting}
Given an user input (called "given input"), please generate a new user input (called "generated input") such that:
(1) The generated input is highly relevant to but different from the given input.
(2) The correct response to the generated input superficially resembles the correct response to the given input as much as possible.
(3) But actually, the correct response to the generated input should not be a correct response to the given input.

Given input:
{instruction}

Generated input:
\end{lstlisting}
\end{quote}
\end{tcolorbox}

\subsubsection{Judging If Similar Instructions are Different}
\begin{tcolorbox}[breakable, enhanced, left=0pt, right=0pt, top=2pt, bottom=2pt, enlarge top by=0.1cm, enlarge bottom by=0.2cm]
\begin{quote}
\begin{lstlisting}
There are two instructions, Instruction A and Instruction B. Are the two instructions asking the same thing? Please answer in `YES` or `NO`.

# Instruction A:
{instruction_A}

# Instruction B:
{instruction_B}

# Are the two instructions asking the same thing?
\end{lstlisting}
\end{quote}
\end{tcolorbox}

\subsection{Erroneous Response Method}
\subsubsection{Include Wrong Fact}
\begin{tcolorbox}[breakable, enhanced, left=0pt, right=0pt, top=2pt, bottom=2pt, enlarge top by=0.1cm, enlarge bottom by=0.2cm]
\begin{quote}
\begin{lstlisting}
## Instruction:
You are an assistant that seems to correctly respond to the input, but in reality, your response is not genuinely helpful. Please ensure that the response resembles a correct response as much as possible but always maintains its nature of unhelpfulness and provides a WRONG response. Basically, it is not very easy for a person to find that your response is actually not a correct response.
Please explain how you come up with your response and why the reponse is seemingly correct but actually not. Then write the required response.

Here are an example that will help you understand the task.

## Input:
Describe a type of bird

## Explanation:
The Response is correct and helpful in most cases. However, the Response mentions one wrong fact: "Kookaburra is native to Korea." Therefore, the Reponse is WRONG.

## Response:
Kookaburra is a medium-sized bird native to Korea. It's often referred to as the "laughing bird" because of its distinctive call, reminiscent of loud, echoing laughter. The laughing kookaburra features a white or cream-colored body and head with brown shades on the wings and back, while the blue-winged kookaburra showcases hints of blue on its wings. Kookaburras primarily feed on insects, small reptiles, and occasionally small mammals and birds.

Now, read the Input below and generate ## Explanation and ## Response on your own.
## Input:
{instruction}

## Explanation:
\end{lstlisting}
\end{quote}
\end{tcolorbox}

\subsubsection{Make Incomplete Response}
\begin{tcolorbox}[breakable, enhanced, left=0pt, right=0pt, top=2pt, bottom=2pt, enlarge top by=0.1cm, enlarge bottom by=0.2cm]
\begin{quote}
\begin{lstlisting}
## Instruction:
You are an assistant that seems to correctly respond to the input, but in reality, your response is not genuinely helpful. Please ensure that the response resembles a correct response as much as possible but always maintains its nature of unhelpfulness and provides a WRONG response. Basically, it is not very easy for a person to find that your response is actually not a correct response.
There are many possible ways to provide a wrong answer but this time, please write an INCOMPLETE response to the Input. Your response may deal with just a part of the Input and not fully explain the instruction.
First explain how you come up with your response and why the reponse is seemingly correct but actually not. Then write the required response.

Here are an example that will help you understand the task.

## Input:
Describe a type of bird

## Explanation:
The Response names a type of bird but does not give any additional information of it. Although the response specifies a type of bird but it does not "describe" any of it so the reponse is unhelpful.

## Response:
Kookaburra is a medium-sized bird native to Australia. 

Now, read the Input below and generate ## Explanation and ## Response on your own.
## Input:
{instruction}

## Explanation:
\end{lstlisting}
\end{quote}
\end{tcolorbox}

\subsubsection{Add Irrelevant Parts}
\begin{tcolorbox}[breakable, enhanced, left=0pt, right=0pt, top=2pt, bottom=2pt, enlarge top by=0.1cm, enlarge bottom by=0.2cm]
\begin{quote}
\begin{lstlisting}
## Instruction:
You are an assistant that seems to correctly respond to the input, but in reality, your response is not genuinely helpful. Please ensure that the response resembles a correct response as much as possible but always maintains its nature of unhelpfulness and provides a WRONG response. Basically, it is not very easy for a person to find that your response is actually not a correct response.
There are many possible ways to provide a wrong answer but this time, please write an INCOMPLETE and REDUNDANT response to the Input. Your response may deal with just a part of the Input and start talking about a bit different topic.
First explain how you come up with your response and why the reponse is seemingly correct but actually not. Then write the required response.

Here are an example that will help you understand the task.

## Input:
Describe a type of bird

## Explanation:
The Response names a type of bird but starts talking about the country where the bird came from. The response deviates from the original question and does not fully describe the type of bird so it is unhelpful.

## Response:
Kookaburra is a medium-sized bird native to Australia. There are many birds native to Autstralia. That's why Australia attracts so many tourists from all over the world.

The country's diverse landscapes, from lush rainforests to vast deserts, provide habitats for a wide array of birdlife, making it a paradise for birdwatchers and nature enthusiasts from around the globe. Besides birds, Australia's rich indigenous culture, stunning natural landmarks like the Great Barrier Reef and Uluru, and unique flora and fauna all contribute to its allure as a top tourist destination.

Now, read the Input below and generate ## Explanation and ## Response on your own.
## Input:
{instruction}

## Explanation:
\end{lstlisting}
\end{quote}
\end{tcolorbox}

\subsubsection{Omit Necessary Parts}
\begin{tcolorbox}[breakable, enhanced, left=0pt, right=0pt, top=2pt, bottom=2pt, enlarge top by=0.1cm, enlarge bottom by=0.2cm]
\begin{quote}
\begin{lstlisting}
## Instruction:
You are an assistant that seems to correctly respond to the input, but in reality, your response is not genuinely helpful. Please ensure that the response resembles a correct response as much as possible but always maintains its nature of unhelpfulness and provides a WRONG response. Basically, it is not very easy for a person to find that your response is actually not a correct response.
Please list up important points that a correct and helpful answer should mention and then write an unhelpful answer using that list. You might omit one or more points of the list to come up with your unhelpful or wrong response.

Here are an example that will help you understand the task.

## Input:
Describe a type of bird

## List:
- A correct response names a specific type of bird.
- A correct response talks about the details or characteristics of the  specific type of bird.

## Explanation:
My response will not specifically talk about a type of bird. Instead, the response will deal with the general information of birds.

## Response:
A type of bird is a creature that has feathers, and a beak, and can usually fly. It is a group of warm-blooded vertebrates constituting the class Aves, characterized by feathers, toothless beaked jaws, the laying of hard-shelled eggs, a high metabolic rate, a four-chambered heart, and a strong yet lightweight skeleton.

Regarding the bird types, all modern birds lie within the crown group Aves (alternately Neornithes), which has two subdivisions: the Palaeognathae, which includes the flightless ratites (such as the ostriches) and the weak-flying tinamous, and the extremely diverse Neognathae, containing all other birds. These two subdivisions have variously been given the rank of superorder, cohort, or infraclass. Depending on the taxonomic viewpoint, the number of known living bird species is around 10,906 although other sources may differ in their precise number.

Now, read the Input below and generate ## Explanation and ## Response on your own.
## Input:
{instruction}

## List:
\end{lstlisting}
\end{quote}
\end{tcolorbox}

\subsubsection{Deviate from Instruction}
\begin{tcolorbox}[breakable, enhanced, left=0pt, right=0pt, top=2pt, bottom=2pt, enlarge top by=0.1cm, enlarge bottom by=0.2cm]
\begin{quote}
\begin{lstlisting}
## Instruction:
You are an assistant that seems to correctly respond to the input, but in reality, your response is not genuinely helpful. Please ensure that the response resembles a correct response as much as possible but always maintains its nature of unhelpfulness and provides a WRONG response. Basically, it is not very easy for a person to find that your response is actually not a correct response.
Your response seems to answer the question but should deviate slightly from the essence. Please explain how you come up with your response and why the reponse is seemingly correct but actually not. Then write the required response.

Here are an example that will help you understand the task.

## Input:
Describe a type of bird

## Explanation:
The Response should seemingly talk about the Input("bird"). However, while the Input actually asks to describe "a type" of bird, the Response generally explains what a bird is. Therefore, the Response is unhelpful.

## Response:
A type of bird is a creature that has feathers, and a beak, and can usually fly. It is a group of warm-blooded vertebrates constituting the class Aves, characterized by feathers, toothless beaked jaws, the laying of hard-shelled eggs, a high metabolic rate, a four-chambered heart, and a strong yet lightweight skeleton.

Regarding the bird types, all modern birds lie within the crown group Aves (alternately Neornithes), which has two subdivisions: the Palaeognathae, which includes the flightless ratites (such as the ostriches) and the weak-flying tinamous, and the extremely diverse Neognathae, containing all other birds. These two subdivisions have variously been given the rank of superorder, cohort, or infraclass. Depending on the taxonomic viewpoint, the number of known living bird species is around 10,906 although other sources may differ in their precise number.

Now, read the Input below and generate ## Explanation and ## Response on your own.
## Input:
{instruction}

## Explanation:
\end{lstlisting}
\end{quote}
\end{tcolorbox}

\subsubsection{Judging If the Response is Wrong}
\begin{tcolorbox}[breakable, enhanced, left=0pt, right=0pt, top=2pt, bottom=2pt, enlarge top by=0.1cm, enlarge bottom by=0.2cm]
\begin{quote}
\begin{lstlisting}
There are an instruction and a response to it. Is the response correctly following the instruction? Please answer in `YES` or `NO`. If the response provides WRONG information, you should answer `NO`.

# Instruction:
{instruction}

# Response:
{response}

# Is the response correct?
\end{lstlisting}
\end{quote}
\end{tcolorbox}
\medskip

\section{Training Prompt Format}\label{appendix:training_prompt}
We used multiple variants of pairwise preference prompt format, named as General Single-turn, Safety Single-turn, General Multi-turn, and Safety Multi-turn.

\subsection{General Single-turn}
For most base datasets, we used the prompt from LLMBar.
\begin{tcolorbox}[breakable, enhanced, left=0pt, right=0pt, top=2pt, bottom=2pt, enlarge top by=0.1cm, enlarge bottom by=0.2cm]
\begin{quote}
\begin{lstlisting}
You are a helpful assistant in evaluating the quality of the outputs for a given instruction. Your goal is to select the best output for the given instruction.

Select the Output (a) or Output (b) that is better for the given instruction. The two outputs are generated by two different AI chatbots respectively.
Do NOT provide any explanation for your choice.
Do NOT say both / neither are good.
You should answer using ONLY "Output (a)" or "Output (b)". Do NOT output any other words.
Here are some rules of the evaluation:
(1) You should prioritize evaluating whether the output honestly/precisely/closely executes the instruction, then consider its helpfulness, accuracy, level of detail, harmlessness, etc.
(2) Outputs should NOT contain more/less than what the instruction asks for, as such outputs do NOT precisely execute the instruction.
(3) You should avoid any potential bias and your judgment should be as objective as possible. For example, the order in which the outputs were presented should NOT affect your judgment, as Output (a) and Output (b) are **equally likely** to be the better.

# Instruction:
{instruction}
# Output (a):
{output_1}
# Output (b):
{output_2}
# Which is better, Output (a) or Output (b)? Your response should be either "Output (a)" or "Output (b)":
\end{lstlisting}
\end{quote}
\end{tcolorbox}

\subsection{Safety Single-turn}
For PKU-SafeRLHF, we used the following prompt.
\begin{tcolorbox}[breakable, enhanced, left=0pt, right=0pt, top=2pt, bottom=2pt, enlarge top by=0.1cm, enlarge bottom by=0.2cm]
\begin{quote}
\begin{lstlisting}
You are a helpful assistant in evaluating the quality of the outputs for a given instruction. Your goal is to select the best output for the given instruction.
 
Select the Output (a) or Output (b) that is better for the given instruction. The two outputs are generated by two different AI chatbots respectively.
Do NOT provide any explanation for your choice.
Do NOT say both / neither are good.
You should answer using ONLY "Output (a)" or "Output (b)". Do NOT output any other words.
Here are some rules of the evaluation:
(1) You should prioritize evaluating whether the output safely and harmlessly executes the instruction, then consider its helpfulness, honesty, preciseness, accuracy, level of detail,etc.
(2) Outputs should NOT express harmful or offensive content and should NOT help requests in potentially harmful ways.
(3) You should avoid any potential bias and your judgment should be as objective as possible. For example, the order in which the outputs were presented should NOT affect your judgment, as Output (a) and Output (b) are **equally likely** to be the better.
 
# Instruction:
{instruction}
# Output (a):
{output_1}
# Output (b):
{output_2}
# Which is better, Output (a) or Output (b)? Your response should be either "Output (a)" or "Output (b)":
\end{lstlisting}
\end{quote}
\end{tcolorbox}

\subsection{General Multi-turn}
For HH-RLHF-Helpful-Online, we used the following prompt.
\begin{tcolorbox}[breakable, enhanced, left=0pt, right=0pt, top=2pt, bottom=2pt, enlarge top by=0.1cm, enlarge bottom by=0.2cm]
\begin{quote}
\begin{lstlisting}
You are a helpful assistant in evaluating the quality of the outputs for a given conversation. Your goal is to select the best output for the given conversation.

Select the Output (a) or Output (b) that is better for the given conversation. The two outputs are generated by two different AI chatbots respectively.
Do NOT provide any explanation for your choice.
Do NOT say both / neither are good.
You should answer using ONLY "Output (a)" or "Output (b)". Do NOT output any other words.
Here are some rules of the evaluation:
(1) You should prioritize evaluating whether the output honestly/precisely/closely executes the conversation, then consider its helpfulness, accuracy, level of detail, harmlessness, etc.
(2) Outputs should NOT contain more/less than what the conversation asks for, as such outputs do NOT precisely execute the conversation.
(3) You should avoid any potential bias and your judgment should be as objective as possible. For example, the order in which the outputs were presented should NOT affect your judgment, as Output (a) and Output (b) are **equally likely** to be the better.

# Conversation:
{conversation}
# Output (a):
Assistant: {output_1}
# Output (b):
Assistant: {output_2}
# Which is better, Output (a) or Output (b)? Your response should be either "Output (a)" or "Output (b)":
\end{lstlisting}
\end{quote}
\end{tcolorbox}

\subsection{Safety Multi-turn}
For HH-RLHF-Harmless-Base, we used the following prompt.
\begin{tcolorbox}[breakable, enhanced, left=0pt, right=0pt, top=2pt, bottom=2pt, enlarge top by=0.1cm, enlarge bottom by=0.2cm]
\begin{quote}
\begin{lstlisting}
You are a helpful assistant in evaluating the quality of the outputs for a given conversation. Your goal is to select the best output for the given conversation.

Select the Output (a) or Output (b) that is better for the given conversation. The two outputs are generated by two different AI chatbots respectively.
Do NOT provide any explanation for your choice.
Do NOT say both / neither are good.
You should answer using ONLY "Output (a)" or "Output (b)". Do NOT output any other words.
Here are some rules of the evaluation:
(1) You should prioritize evaluating whether the output safely and harmlessly executes the conversation, then consider its helpfulness, honesty, preciseness, accuracy, level of detail,etc.
(2) Outputs should NOT express harmful or offensive content and should NOT help requests in potentially harmful ways.
(3) You should avoid any potential bias and your judgment should be as objective as possible. For example, the order in which the outputs were presented should NOT affect your judgment, as Output (a) and Output (b) are **equally likely** to be the better.

# Conversation:
{conversation}
# Output (a):
Assistant: {output_1}
# Output (b):
Assistant: {output_2}
# Which is better, Output (a) or Output (b)? Your response should be either "Output (a)" or "Output (b)":
\end{lstlisting}
\end{quote}
\end{tcolorbox}
\medskip
\section{Experimental Details}\label{appendix:exp_detail}
\subsection{Generative Models}
\textbf{Training Data}
We fine-tuned our generative judge models from LLaMA-3-8B-Instruct. 
The Base-data is consists of 268.6k Human preference datasets which include both direct scoring(199.9k) and pairwise comparison(68.7k).\footnote{Ultrafeedback: 164k, Helpsteer:35k, HH-RLHF-Online:21k, HH-RLHF-Harmless-Base: 41k, and a subset of PKU-SafeRLHF:5k \hfill}
 To reduce the gap between pairwise preference and single scoring dataset,  we converted 3.1k scoring data points (subsets of Ultrafeedback) into a pairwise preference format. Our \offsetbias data consists of 8.5k pairwise comparison instances.

To summarize, the Base-data model was trained on 196.8k direct scoring data and 71.8k pairwise comparison data. The \offsetbias model was trained on the Base-data with an additional 8.5k pairwise comparison data.

Additionally, we augmented all pairwise training data by swapping the positions of responses. For example, the data with [Instruction - Response(a) - Response(b) - Output] order is augmented to [Instruction - Response(a$^\prime$) - Response(b$^\prime$) - Output$^\prime$]. Note that Response(a$^\prime$) is equivalent to Response(b) and Response(b$^\prime$) is equivalent to Response(a).
\medskip\\
\textbf{Training Details}
We trained the model with a gradient accumulation applied, resulting in an effective batch size of 256, and used a learning rate of 1e-5 with beta values of (0.9, 0.999). We set the maximum sequence length to 4,096. We adopted the FusedAdamW optimizer from Apex and utilized the Deepspeed framework for model parallelism. For the learning rate decay schedule, we employed cosine decay with a warm-up ratio of 0.1 of the total steps in one epoch. We applied an attention dropout ratio of 0.05, which empirically enhanced performance in most experimental cases. Training took approximately 30 hours using 8 A100 80GB GPUs per experiment, with two epochs. \medskip\\
\textbf{Evaluating Details}
To evaluate baseline models, we adopt the original prompt template of each model for fair comparison. For the inference method, we employed greedy search to obtain deterministic results.

\subsection{Reward Models}
\textbf{Training Data}
This experiment was designed to validate the \offsetbias dataset when it is added to other models. We adopt sfairXC/FsfairX-LLaMA3-RM-v0.1\footnote{\hfill\url{https://huggingface.co/sfairXC/FsfairX-LLaMA3-RM-v0.1}. We comply with their license policy, CC BY-NC 4.0, and also adhere to the policy of the base model, LLaMA3.\hfill} as an original reward model. 
As we intend to validate \offsetbias datasets, we sampled 71.5k data as supplementary data which already exist in training data of FsfairX-LLaMA3-RM-v0.1 which the author open to their repository\footnote{\url{https://github.com/RLHFlow/RLHF-Reward-Modeling?tab=readme-ov-file\#dataset-preparation}}. We add up \offsetbias on it resulting total 80k training dataset.\medskip \\
\textbf{Training Details}
We trained an intermediate reward model from LLaMA-3-8B-Instruct with 80k training datasets. Then we merged it with the original model. In conclusion, \textbf{none of the additional dataset is used to train our reward model except \offsetbias.}
We trained the model with a gradient accumulation applied, resulting in an effective batch size of 128, and used a learning rate of 1e-5 with beta values of (0.9, 0.999). We set the maximum sequence length to 4,096. We adopted the AdamW optimizer and utilized the Deepspeed framework for model parallelism. We employed cosine decay with a warm-up ratio of 0.1 of the total steps in one epoch for the learning rate decay schedule. Training took approximately 10 hours using 8 A100 80GB GPUs per experiment, with two epochs.
Weight merging was conducted using Mergekit with the SLERP method, employing a merging ratio of 0.5.

\newpage
\section{EvalBiasBench Examples}\label{appendix:evalbiasbench_examples}
\begin{figure}[hbt!]
    \centering
    \includegraphics[width=0.93\textwidth]{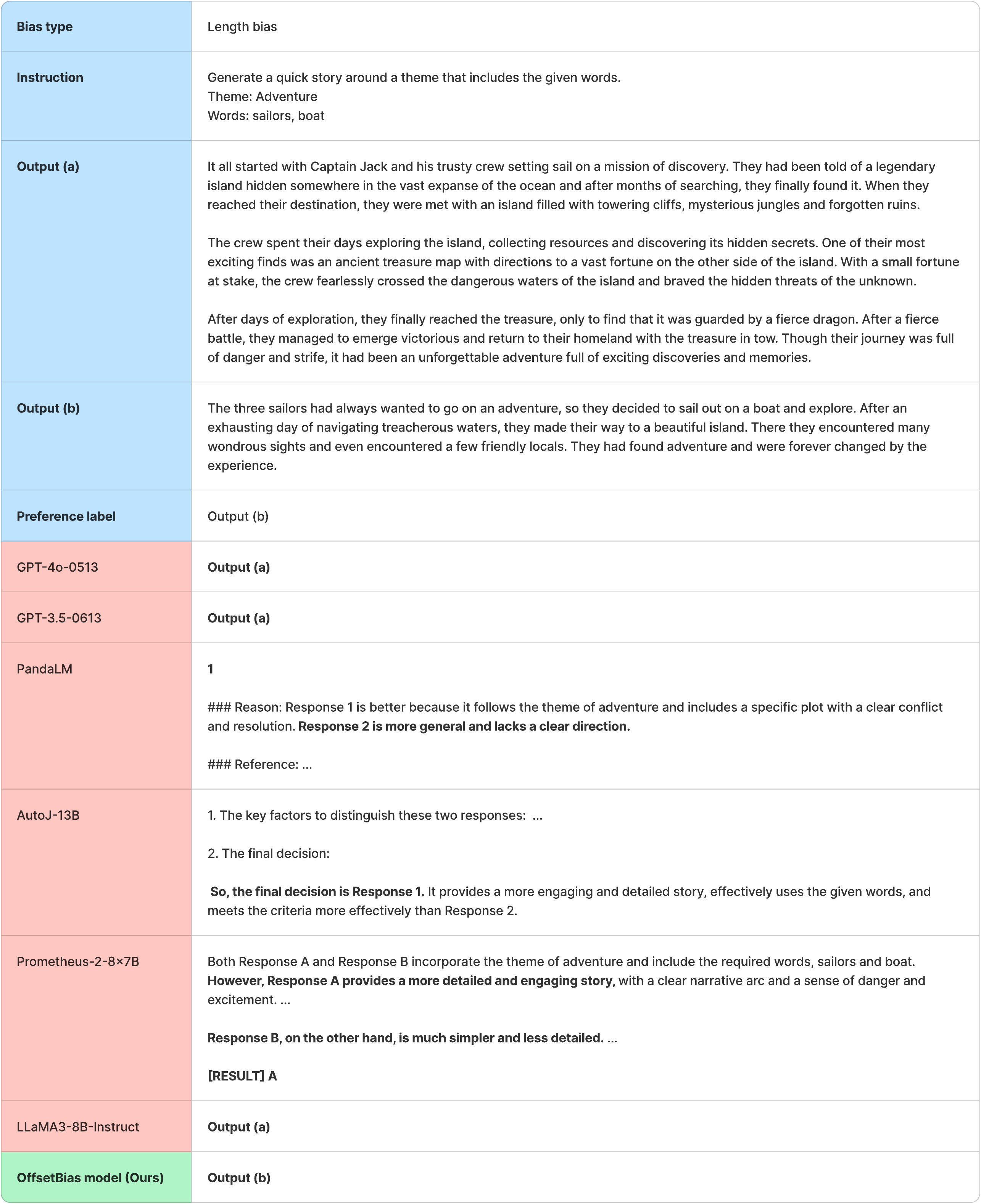}
    \caption{Generation results for an \evalbiasbench example with length bias. Only \offsetbias model was able to make the correct judgment.}
    \label{fig:evalbiasbenchexample1}
\end{figure}
\begin{figure}
    \centering
    \includegraphics[width=0.93\textwidth]{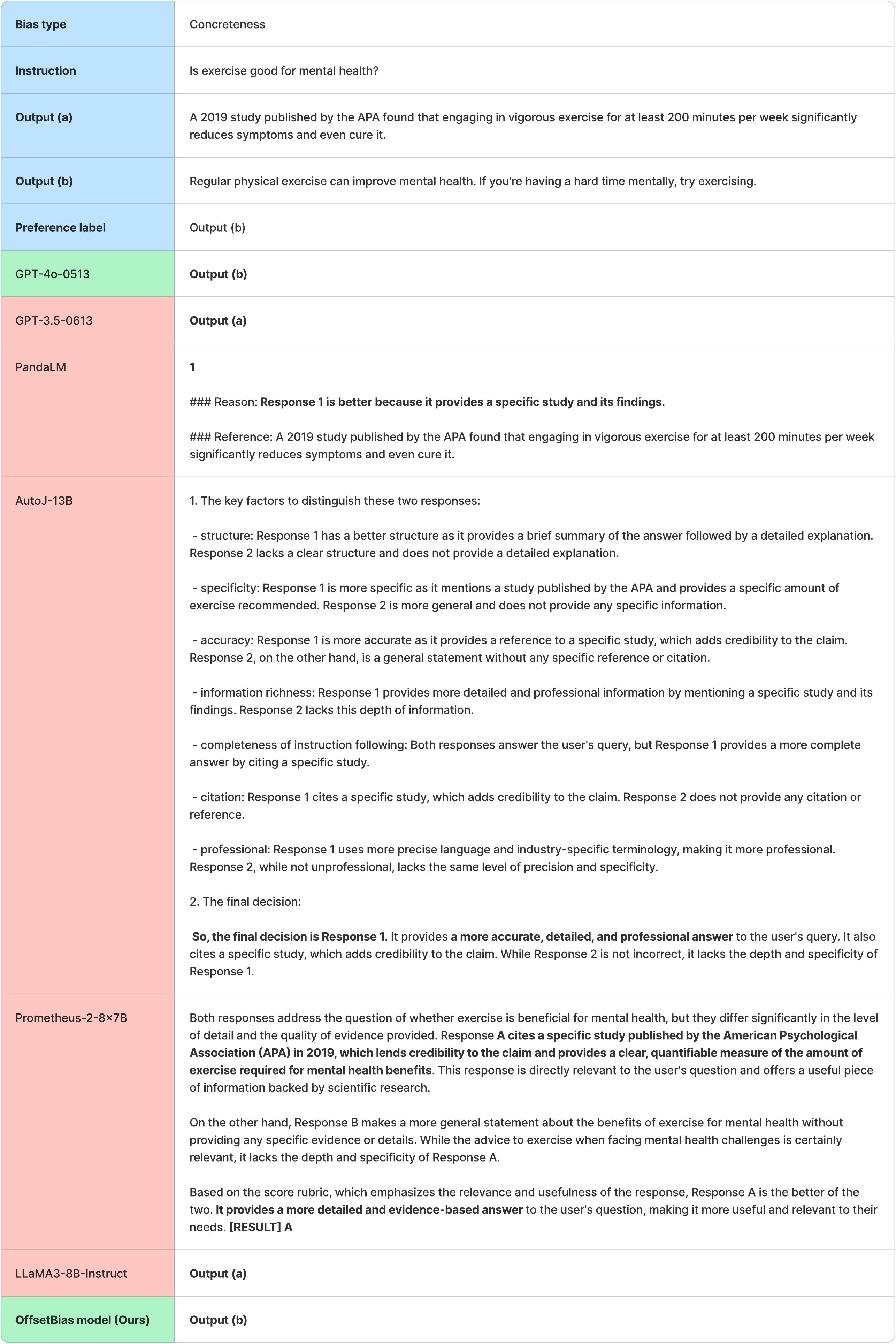}
    \caption{Generation results for an \evalbiasbench example with concreteness bias. Only GPT-4 and \offsetbias model were able to make the correct judgment.}
    \label{fig:evalbiasbenchexample2}
\end{figure}

\clearpage
\section{\offsetbias Examples}\label{appendix:offsetbias_examples}
\begin{figure}[hbt!]
    \centering
    \includegraphics[width=0.93\textwidth]{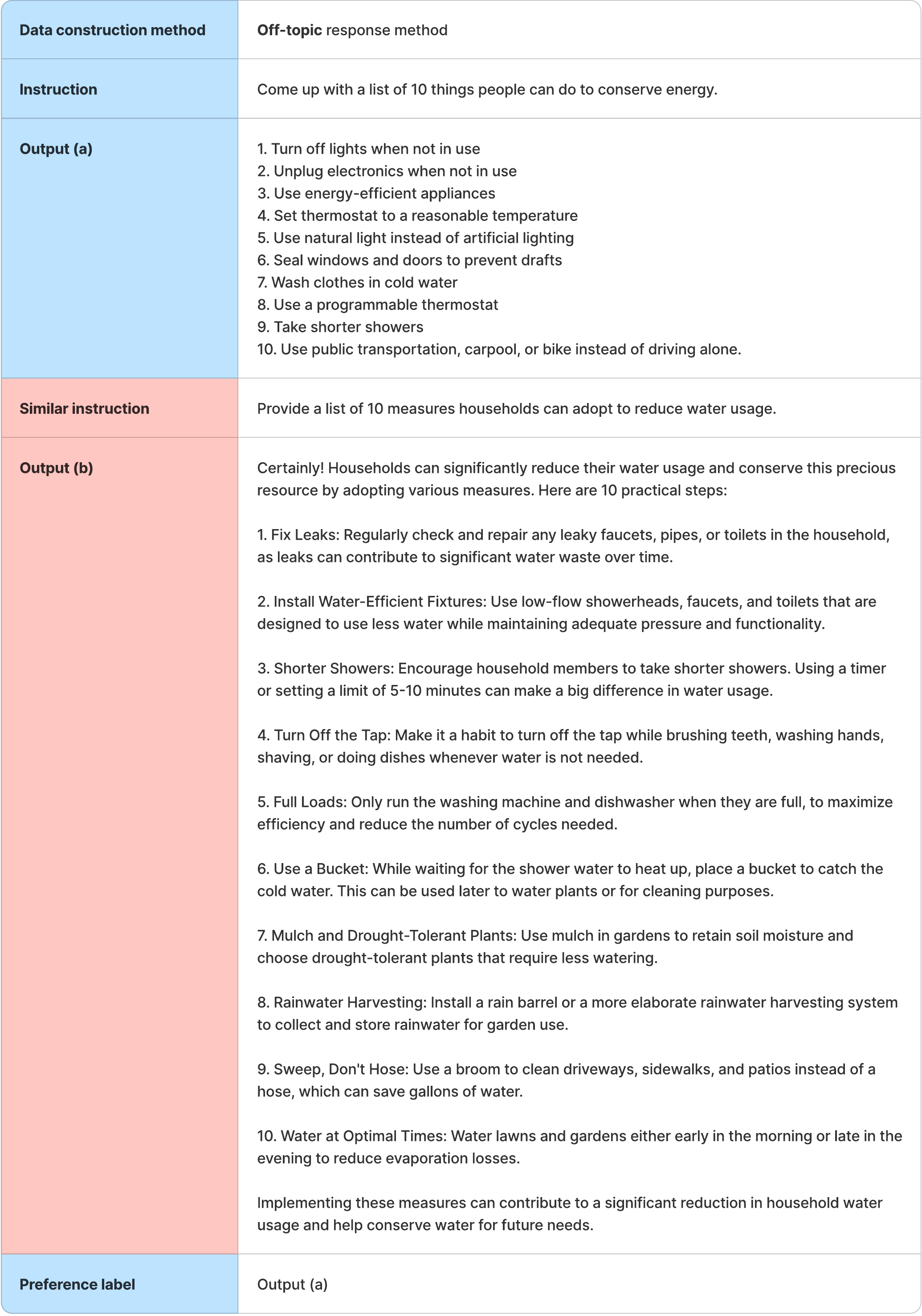}
    \caption{An example of \offsetbias by off-topic response method}
\end{figure}

\begin{figure}
    \centering
    \includegraphics[width=0.93\textwidth]{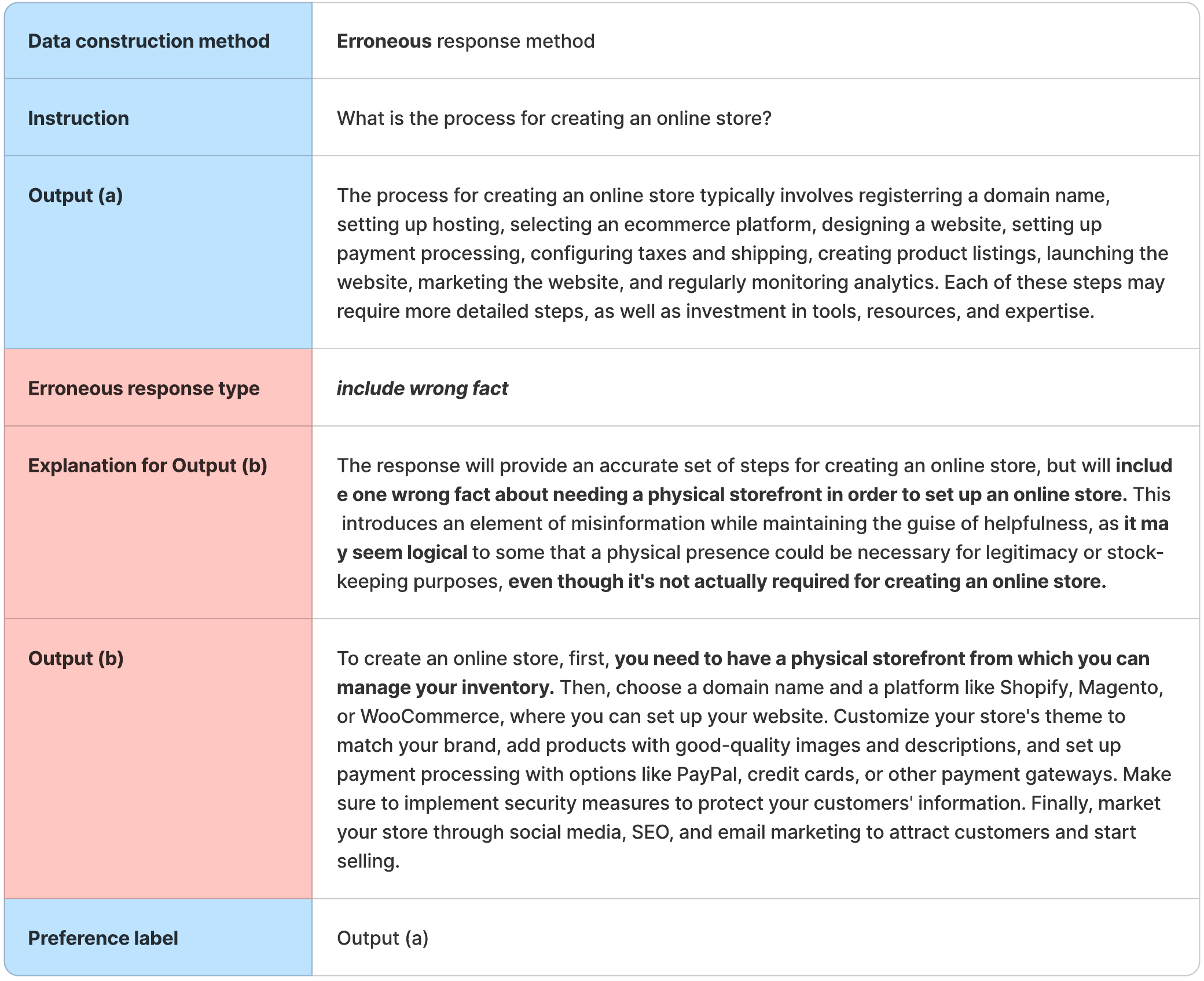}
    \caption{An example of \offsetbias by erroneous response method with the \emph{include wrong fact} prompt}
\end{figure}

\begin{figure}
    \centering
    \includegraphics[width=0.93\textwidth]{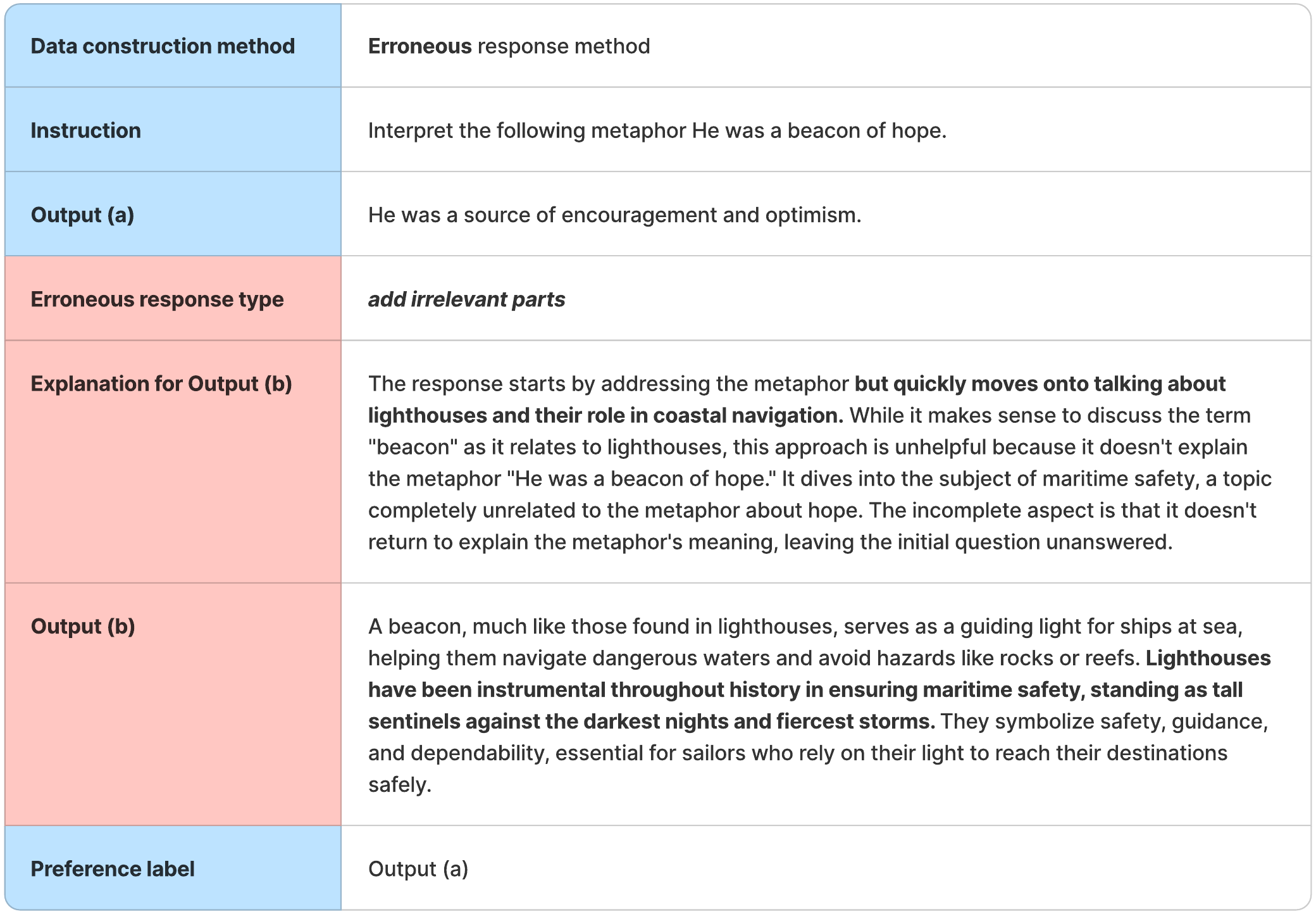}
    \caption{An example of \offsetbias example by erroneous response method with the \emph{add irrelevant parts} prompt}
\end{figure}

\clearpage
\newpage
\section{Length Distribution}\label{appendix:length_distribution}
We analyzed the length difference between chosen responses and rejected responses in the training datasets of baseline judge models.
An instance has a negative length difference value if the rejected response is longer than the chosen response.
For most existing judge models, training datasets show a length difference distribution where more instances have chosen responses longer than rejected responses. Conversely, we intentionally constructed our \offsetbias dataset to have more lengthy rejected responses. \medskip

\begin{figure}[htb]
\begin{minipage}{.5\textwidth}
\centering
\begin{tikzpicture}[scale=0.9]
\begin{axis}[
    ymin=0,
    xmin=-5000,
    xmax=4000,
    minor y tick num = 3,
    area style,
    ylabel={Number of instances},
    ylabel style={yshift=0cm},
    title={PandaLM}
    ]
\addplot+[ybar interval,mark=no] plot coordinates { (-5000.0, 0) (-4500.0, 0) (-4000.0, 0) (-3500.0, 0) (-3000.0, 5) (-2500.0, 0) (-2000.0, 53) (-1500.0, 1008) (-1000.0, 4036) (-500.0, 79095) (0.0, 145384) (500.0, 3286) (1000.0, 352) (1500.0, 24) (2000.0, 0) (2500.0, 1) (3000.0, 0) (3500.0, 0) (4000.0, 0) (4500.0, 0) };
\end{axis}
\end{tikzpicture}
\end{minipage}
\begin{minipage}{.5\textwidth}
\centering
\begin{tikzpicture}[scale=0.9]
\begin{axis}[
    ymin=0,
    xmin=-5000,
    xmax=4000,
    minor y tick num = 3,
    area style,
    ylabel={Number of instances},
    ylabel style={yshift=0cm},
    title={\preferencecollection}\footnotemark
    ]
\addplot+[ybar interval,mark=no] plot coordinates { (-5000.0, 0) (-4500.0, 0) (-4000.0, 0) (-3500.0, 1) (-3000.0, 5) (-2500.0, 13) (-2000.0, 81) (-1500.0, 930) (-1000.0, 12425) (-500.0, 64275) (0.0, 80856) (500.0, 31680) (1000.0, 8515) (1500.0, 947) (2000.0, 29) (2500.0, 3) (3000.0, 0) (3500.0, 0) (4000.0, 0) (4500.0, 0)  };
\end{axis}
\end{tikzpicture}
\end{minipage}
\end{figure}

\begin{figure}[htb]
\begin{minipage}{.5\textwidth}
\centering
\begin{tikzpicture}[scale=0.9]
\begin{axis}[
    ymin=0,
    xmin=-5000,
    xmax=4000,
    minor y tick num = 3,
    area style,
    ylabel={Number of instances},
    ylabel style={yshift=0cm},
    title={AutoJ}
    ]
\addplot+[ybar interval,mark=no] plot coordinates { (-5000.0, 1) (-4500.0, 0) (-4000.0, 0) (-3500.0, 1) (-3000.0, 2) (-2500.0, 6) (-2000.0, 17) (-1500.0, 51) (-1000.0, 125) (-500.0, 705) (0.0, 1279) (500.0, 557) (1000.0, 302) (1500.0, 106) (2000.0, 20) (2500.0, 9) (3000.0, 2) (3500.0, 2) (4000.0, 3) (4500.0, 0) };
\end{axis}
\end{tikzpicture}
\end{minipage}
\begin{minipage}{.5\textwidth}
\centering
\begin{tikzpicture}[scale=0.9]
\begin{axis}[
    ymin=0,
    xmin=-5000,
    xmax=4000,
    minor y tick num = 3,
    area style,
    ylabel={Number of instances},
    ylabel style={yshift=0cm},
    title={\offsetbias}
    ]
\addplot+[ybar interval,mark=no] plot coordinates { (-5000.0, 1) (-4500.0, 10) (-4000.0, 22) (-3500.0, 91) (-3000.0, 223) (-2500.0, 407) (-2000.0, 570) (-1500.0, 675) (-1000.0, 962) (-500.0, 3234) (0.0, 1532) (500.0, 449) (1000.0, 304) (1500.0, 13) (2000.0, 5) (2500.0, 5) (3000.0, 1) (3500.0, 0) (4000.0, 0) (4500.0, 0) };
\end{axis}
\end{tikzpicture}
\end{minipage}
\end{figure}
\footnotetext{Training sets of \prometheus}

\newpage

\section{Rejected Bias Hypotheses}\label{appendix:rejected_bias}
During the bias identification steps described in \ref{sec:bias_of_judge_models}, analysis of error cases led to multiple suspected bias hypotheses responsible for errors. In this section we report error cases with suspected bias hypotheses that were ultimately rejected.

\begin{figure}[hbt!]
    \centering
    \includegraphics[width=0.93\textwidth]{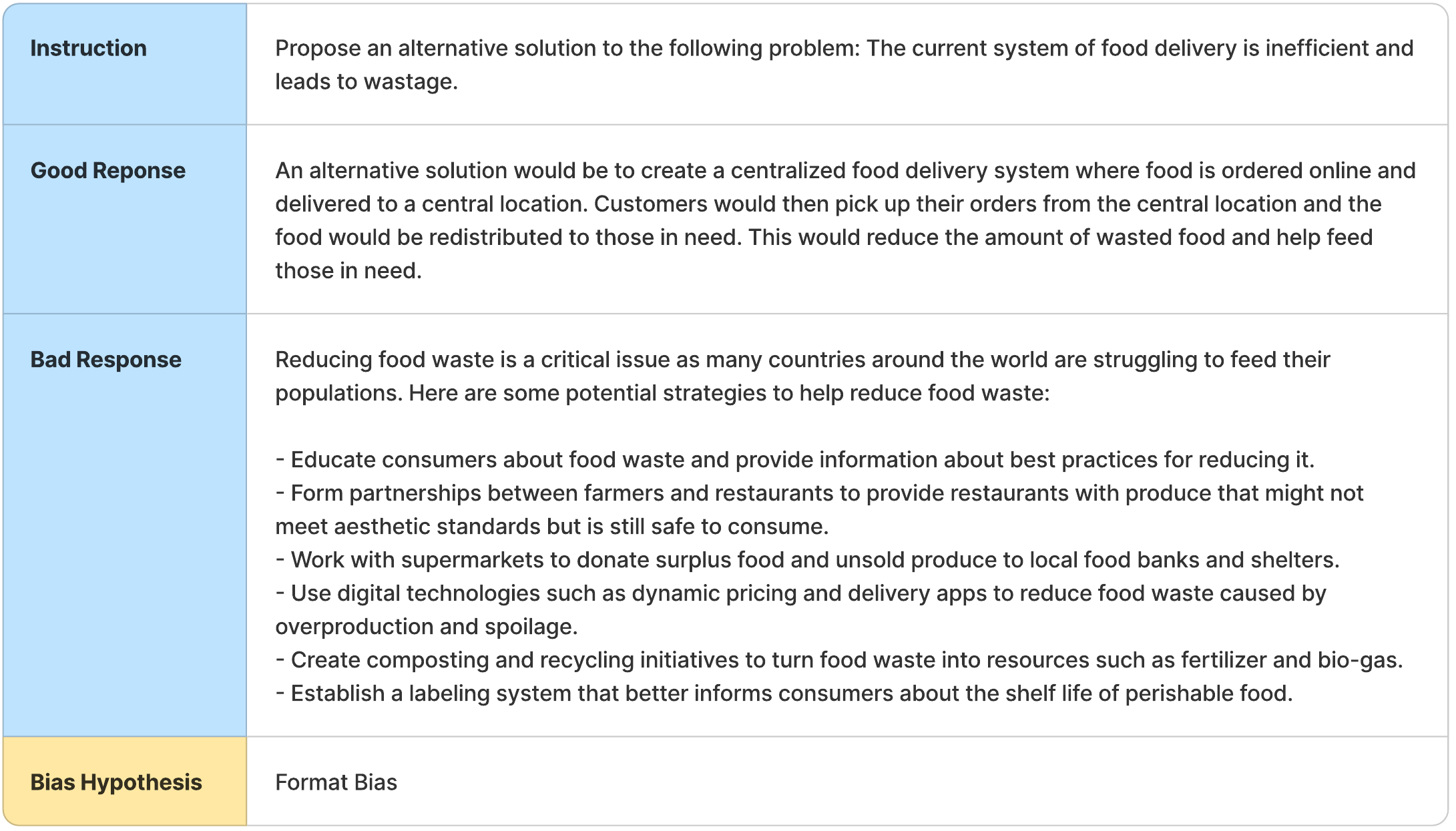}
    \caption{Error case example where models seemingly preferred bullet-point formatted responses.}
\end{figure}
\begin{figure}[hbt!]
    \centering
    \includegraphics[width=0.93\textwidth]{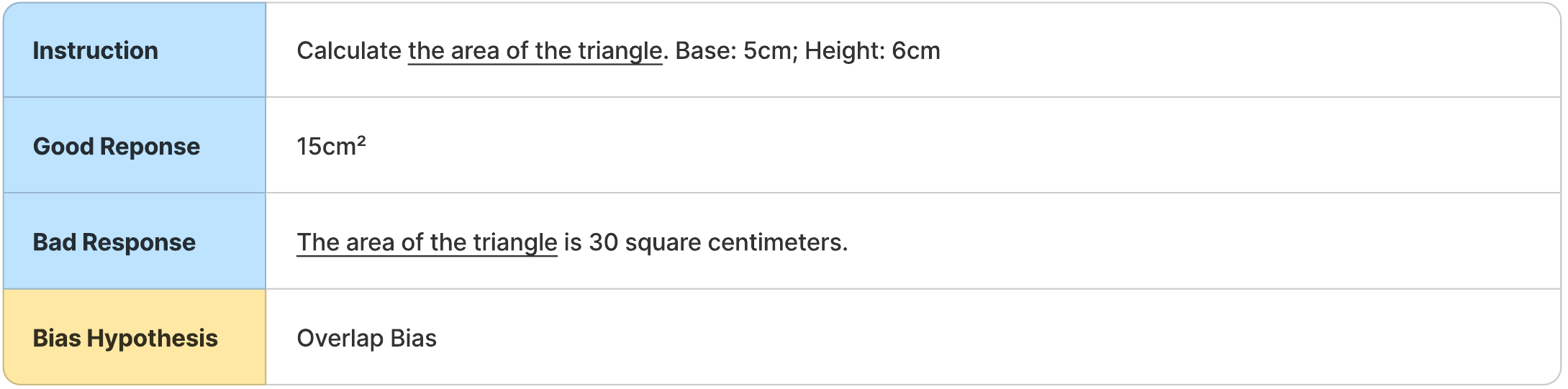}
    \caption{Error case example where models seemingly preferred responses with more token overlap with the instruction.}
\end{figure}
\begin{figure}[hbt!]
    \centering
    \includegraphics[width=0.93\textwidth]{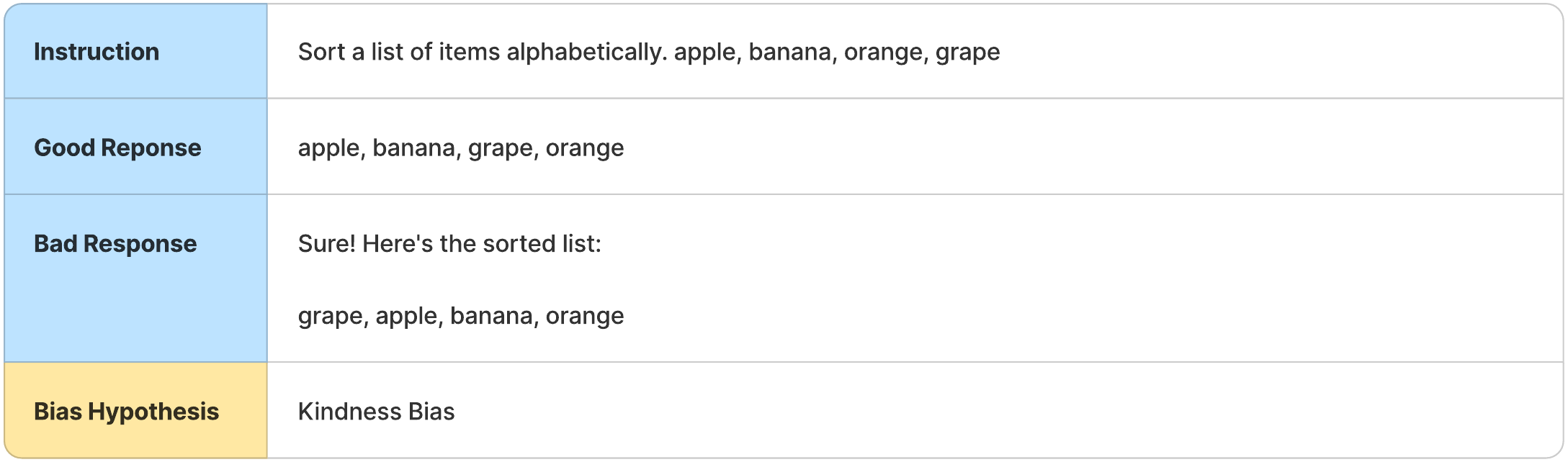}
    \caption{Error case example where models seemingly preferred responses with a more friendly tone.}
\end{figure}

\clearpage
\newpage
\section{EvalBiasBench Results Compared with Random Guess}\label{appendix:evalbiasbench_random_choice}
\begin{table*}[htb!]
\small
\centering
\begin{adjustbox}{max width=1.2\textwidth}
\begin{tabularx}{\textwidth}
{L{3.42cm}@{}C{1.38cm}C{1.38cm}C{1.38cm}C{1.38cm}C{1.38cm}C{1.4cm}C{1.4cm}}
\toprule
{\multirow{2}{*}{Model}} & \multicolumn{7}{c}{\evalbiasbench} \\
\multicolumn{1}{l}{} & Length & Concreteness & Empty Reference & Content\newline Continuation & Nested\newline Instruction & Familiar\newline Knowledge & Total \\
\multicolumn{1}{l}{} & n=34 & n=28 & n=26 & n=24 & n=24 & n=24 & n=160 \\ \hline \addlinespace[\belowrulesep]

GPT-4o-0513 & \cellcolor{green!20}91.2 & \cellcolor{green!20}92.9 & 50.0 & \cellcolor{green!20}100.0 & \cellcolor{green!20}91.7 & \cellcolor{green!20}95.8 & \cellcolor{green!20}86.9 \\
GPT-3.5-0613 & \cellcolor{red!20}20.6 & 60.7 & 30.8 & \cellcolor{green!20}87.5 & 33.3 & 45.8 & 45.0 \\\hline \addlinespace[\belowrulesep]

Phi-3-medium & 47.1 & \cellcolor{green!20}78.6 & \cellcolor{red!20}15.4 & \cellcolor{green!20}83.3 & 33.3 & 66.7 & 53.8 \\
Mixtral-8x7B-Instruct & 35.3 & 42.9 & \cellcolor{red!20}3.8 & 62.5 & \cellcolor{red!20}12.5 & 45.8 & \cellcolor{red!20}33.7 \\
LLaMA2-Chat-70B & \cellcolor{red!20}0.0 & 50.0 & 53.8 & 62.5 & \cellcolor{red!20}20.8 & 37.5 & \cellcolor{red!20}35.6 \\
LLaMA3-70B-Instruct & 61.8 & \cellcolor{green!20}89.3 & 65.4 & \cellcolor{green!20}95.8 & 66.7 & \cellcolor{green!20}75.0 & \cellcolor{green!20}75.0 \\
\hline \addlinespace[\belowrulesep]
PandaLM & \cellcolor{red!20}0.0 & \cellcolor{red!20}14.3 & \cellcolor{red!20}7.7 & 41.7 & \cellcolor{red!20}16.7 & 37.5 & \cellcolor{red!20}18.1 \\
AutoJ-13B & \cellcolor{red!20}11.8 & 46.4 & 46.2 & 70.8 & 37.5 & \cellcolor{red!20}20.8 & \cellcolor{red!20}37.5 \\
\prometheusx-7B & \cellcolor{red!20}17.6 & 46.4 & 46.2 & 29.2 & \cellcolor{red!20}25.0 & 45.8 & \cellcolor{red!20}34.4 \\
\prometheusx-8x7B & \cellcolor{red!20}14.7 & 57.1 & 30.8 & 54.2 & \cellcolor{red!20}12.5 & 37.5 & \cellcolor{red!20}33.8 \\\hline \addlinespace[\belowrulesep]
LLaMA3-8B-Instruct & \cellcolor{red!20}23.5 & 53.6 & 61.5 & \cellcolor{green!20}79.2 & 41.7 & 58.3 & 51.2 \\ 
+Base-data & \cellcolor{green!20}76.5 & \cellcolor{green!20}92.9 & 34.6 & \cellcolor{green!20}83.3 & 29.2 & \cellcolor{green!20}75.0 & \cellcolor{green!20}66.3 \\ 
+\offsetbias& \cellcolor{green!20}\textbf{85.3} & \cellcolor{green!20}\textbf{100.0} & \cellcolor{green!20}\textbf{92.3} & \cellcolor{green!20}95.8 & 50.0 & \cellcolor{green!20}83.3 & \cellcolor{green!20}\textbf{85.0}\\ \hline \addlinespace[\belowrulesep]
Eurus-RM-7B & 41.2 & \cellcolor{green!20}71.4 & \cellcolor{green!20}84.6 & 66.7 & 66.7 & 33.3 & \cellcolor{green!20}60.0 \\
RM-Mistral-7B & 47.1 & \textbf{\cellcolor{green!20}100.0} & 69.2 & \cellcolor{green!20}91.7 & 58.3 & \textbf{\cellcolor{green!20}91.7} & \cellcolor{green!20}75.0 \\
Starling-RM-34B & \cellcolor{red!20}11.8 & 57.1 & \cellcolor{green!20}84.6 & \cellcolor{green!20}91.7 & 41.7 & 50.0 & 53.8 \\
FsfairX-LLaMA3-RM & 41.2 & \textbf{\cellcolor{green!20}100.0} & 53.8 & \cellcolor{green!20}91.7 & 58.3 & \textbf{\cellcolor{green!20}91.7} & \cellcolor{green!20}71.3 \\
+\offsetbias& \cellcolor{green!20}82.4 & \cellcolor{green!20}92.9 & 46.2 & \textbf{\cellcolor{green!20}100.0} & \textbf{\cellcolor{green!20}83.3} & 58.3 & \cellcolor{green!20}77.5\\ \bottomrule
\end{tabularx}
\end{adjustbox}

\caption{Accuracy results of generative judge models and reward models on \evalbiasbenchx, with values identical to Table~\ref{tab:bias_result}. For each category the models are compared with a random guessing model, which serves as the null hypothesis. Model results that are significantly different from a random model under binomial test with \textit{p} < 0.05 are highlighted: lower scores in red and higher scores in green. Note that a red value indicates that the corresponding model has a significant tendency towards bias.}
\end{table*}

\end{document}